\documentclass{article}

\usepackage{arxiv}

\usepackage[utf8]{inputenc} 
\usepackage[T1]{fontenc}    
\usepackage{hyperref}       
\usepackage{url}            
\usepackage{booktabs}       
\usepackage{amsfonts}       
\usepackage{nicefrac}       
\usepackage{microtype}      
\usepackage{graphicx}

\usepackage{natbib}
\usepackage{doi}
\usepackage{tabularx}
\usepackage[svgnames]{xcolor}
\usepackage{soul}

\usepackage{tabto}

\usepackage{listings}
\usepackage{color}

\definecolor{dkgreen}{rgb}{0,0.6,0}
\definecolor{gray}{rgb}{0.5,0.5,0.5}
\definecolor{mauve}{rgb}{0.58,0,0.82}

\newenvironment{note}  
{\par\color{DarkBlue}\begin{quote}\textbf{Note:\ }}
{\end{quote}\par}

\newenvironment{example}  
{\par\color{Black}\begin{quote}\textbf{Example:\ }}
{\end{quote}\par}

\newenvironment{systemprompt}  
{\par\color{Black}\begin{quote}\textbf{System prompt:\ }}
{\end{quote}\par}

\newenvironment{warning}  
{\par\color{Crimson}\begin{quote}\textbf{Warning:\ }}
{\end{quote}\par}

\title{Integrating curation into scientific publishing to train AI models}

\author{
    \href{https://orcid.org/0000-0002-0211-6416}{\includegraphics[scale=0.06]{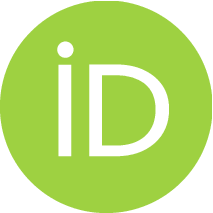}\hspace{1mm}Jorge Abreu Vicente}\\
    Open Science Implementation\\
    EMBO\\
    \And
    \href{https://orcid.org/0000-0001-5461-3265}{\includegraphics[scale=0.06]{orcid.eps}\hspace{1mm}Hannah Sonntag}\\
    Open Science Implementation\\
    EMBO\\
    \And
    \href{https://orcid.org/0000-0002-5472-6355}{\includegraphics[scale=0.06]{orcid.eps}\hspace{1mm}Cassie S. Mitchell}\\
    Department of Biomedical Engineering\\ Georgia Institute of Technology and Emory University School of Medicine\\
    \And
    \href{https://orcid.org/0009-0004-8874-3534}{\includegraphics[scale=0.06]{orcid.eps}\hspace{1mm}Thomas Eidens}\\
    Open Science Implementation\\
    EMBO\\
    \And
    \href{https://orcid.org/0000-0002-2499-4025}{\includegraphics[scale=0.06]{orcid.eps}\hspace{1mm}Thomas Lemberger}\thanks{Corresponding author} \\
    Open Science Implementation\\
    EMBO, Meyerhofstrasse 1, 69117 Heidelberg, Germany\\
    \texttt{thomas.lemberger@embo.org}\\
}

\date{}

\renewcommand{\headeright}{Abreu et al, 2024}
\renewcommand{\undertitle}{}
\renewcommand{\shorttitle}{SourceData-NLP}

\hypersetup{
pdftitle={The SourceData-NLP dataset: integrating curation into scientific publishing for training large language models},
pdfsubject={cs.CL},
pdfauthor={Abreu-J, Sonntag-H, Eidens-T, Lemberger-T},
pdfkeywords={NER, Large Language Models, biocuration},
}

\begin{document}
\maketitle

\begin{abstract}

    
High throughput extraction and structured labeling of data from academic articles is critical to enable downstream machine learning applications and secondary analyses. We have embedded multimodal data curation into the academic publishing process to annotate segmented figure panels and captions. Natural language processing (NLP) was combined with human-in-the-loop feedback from the original authors to increase annotation accuracy. Annotation included eight classes of bioentities (small molecules, gene products, subcellular components, cell lines, cell types, tissues, organisms, and diseases) plus additional classes delineating the entities' roles in experiment designs and methodologies. The resultant dataset, SourceData-NLP, contains more than 620,000 annotated biomedical entities, curated from 18,689 figures in 3,223 articles in molecular and cell biology. We evaluate the utility of the dataset to train AI models using named-entity recognition, segmentation of figure captions into their constituent panels, and a novel context-dependent semantic task assessing whether an entity is a controlled intervention target or a measurement object. We also illustrate the use of our dataset in performing a multi-modal task for segmenting figures into panel images and their corresponding captions.

\end{abstract}

\clearpage

\section{Introduction}
Scientific productivity has reached unprecedented levels. PubMed~\citep{pubmedhelp} alone has over 34 million articles.  Furthermore, millions of new articles are published every year in PubMed and preprint platforms such as  bioRxiv\footnote{https://www.biorxiv.org/}~\citep{biorxiv}. This productivity comes at a cost: the vast amount of literature makes it increasingly difficult for scientists to keep up with the latest research in their field. The "curse of specialization" further exacerbates this problem as scientists become increasingly siloed in their respective areas of expertise. 

To address these challenges, an important goal is the integration of scientific data across the literature into a usable dataset~\citep{sourcedata}. Here we present a powerful approach that embeds the extraction of multimodal data from text, figures, and figure captions in the academic publication process. The integration of biocuration into the academic publishing process enables the capture and exposure of structured, usable datasets as soon as papers are published. It also improves the reliability and rigor of reporting scientific findings by resolving potential ambiguities in terminologies or unclear concepts. Finally, the active involvement and expertise of the authors in the curation process enhances the quality of the captured data for subsequent integrative analyses and artificial intelligence models in biomedical research and predictive medicine.


Natural language processing (NLP), a subset of artificial intelligence methods, enables the computational processing of large volumes of literature. Specifically, named-entity recognition~\citep[NER,][]{bikel-etal-1997-nymble} and named-entity linking~\citep[NEL,][]{hoffart-etal-2011-robust} are important tasks for downstream analysis. NER identifies and categorizes entities within a text, while NEL assigns ontology links or standard identifiers to the entities. These NLP tasks enable the extraction of structured information from scientific literature and facilitate the identification of key concepts and relationships in the data.

The field of biocuration has made significant efforts to assemble large annotated datasets \citep[e.g.,][]{perera,zhao20-biomedical-curation,bigbio}. However, such datasets remain relatively small and specialized. Moreover,  biological datasets curated have traditionally focused on abstracts or selected sentences out of the article general context. However, in the life sciences, the experimental results are mainly presented in figures, with captions providing detailed natural language technical descriptions of the respective experiments~\citep{Li2021UtilizingIA}. As such, figures are the foundation from which the evidence supporting the claims of a paper are presented. Thus, the curation of data contained within figures and figure captions is critical for downstream utilization, including information retrieval and secondary analyses.

Here we present a first-of-its-kind large multimodal dataset, SourceData-NLP, that is constructed by embedding the curation of scientific figures into the academic publishing process. The dataset pairs the images of figure panels with their respective caption segments. The caption text is annotated based on the SourceData entity tagging framework~\citep{sourcedata}, including feedback from the original authors.  The entities extracted represent biological objects across various scales of biological organization -- from small molecules to species -- making the dataset multi-scale in nature.

A unique feature of SourceData-NLP is its ability to capture the essence of the design of perturbation-based experiments by distinguishing between entities that are the objects of the reported measurements and those that are the targets of experimental interventions. This distinction is crucial for understanding  whether a causal hypothesis has been tested in the experiments, setting SourceData-NLP apart from other datasets.

The resulting dataset, SourceData-NLP, provides an integrated, searchable, and ready-to-use database for secondary analyses and tool development. To highlight the utility of SourceData-NLP, we conduct a systematic analysis of the performance of the two best-performing transformers-based~\citep{transformers} models for biomedical NER (BioLinkBERT,~\citealp{biolinkbert}; PubmedBERT,~\citealp{pubmedbert}). Additionally, we demonstrate the utility of our dataset to build object detection-based figure segmentation models that separate scientific figures into their constituent panels and match them to their corresponding part of the figure caption. To facilitate access for researchers to SourceData-NLP, we provide the source code to raw data* in XML and graph database (neo4j) formats\footnote{https://github.com/source-data/soda-data}, machine-learning pre-tokenized data\footnote{https://huggingface.co/datasets/EMBO/SourceData}, the trained models\footnote{https://github.com/source-data/soda-model}, and the multimodal data and models \footnote{https://github.com/source-data/soda\_image\_segmentation}.

\section{Results}

In this section, we provide a comprehensive overview of the SourceData-NLP dataset, emphasizing its key characteristics. We then discuss the outcomes of our empirical analyses to highlight the broad utility of SourceData-NLP. Initially, we present the results for NER benchmarking. Additionally, we introduce a novel semantic task aimed at interpreting the empirical roles of bioentities within experimental setups. Finally, we extend our analysis to include scientific figure segmentation and panel-caption text matching, using an ensemble of object detection and generative AI models.

\subsection{Dataset summary}

\begin{figure}[ht]
    \centering
    \includegraphics[width=1.0\linewidth]{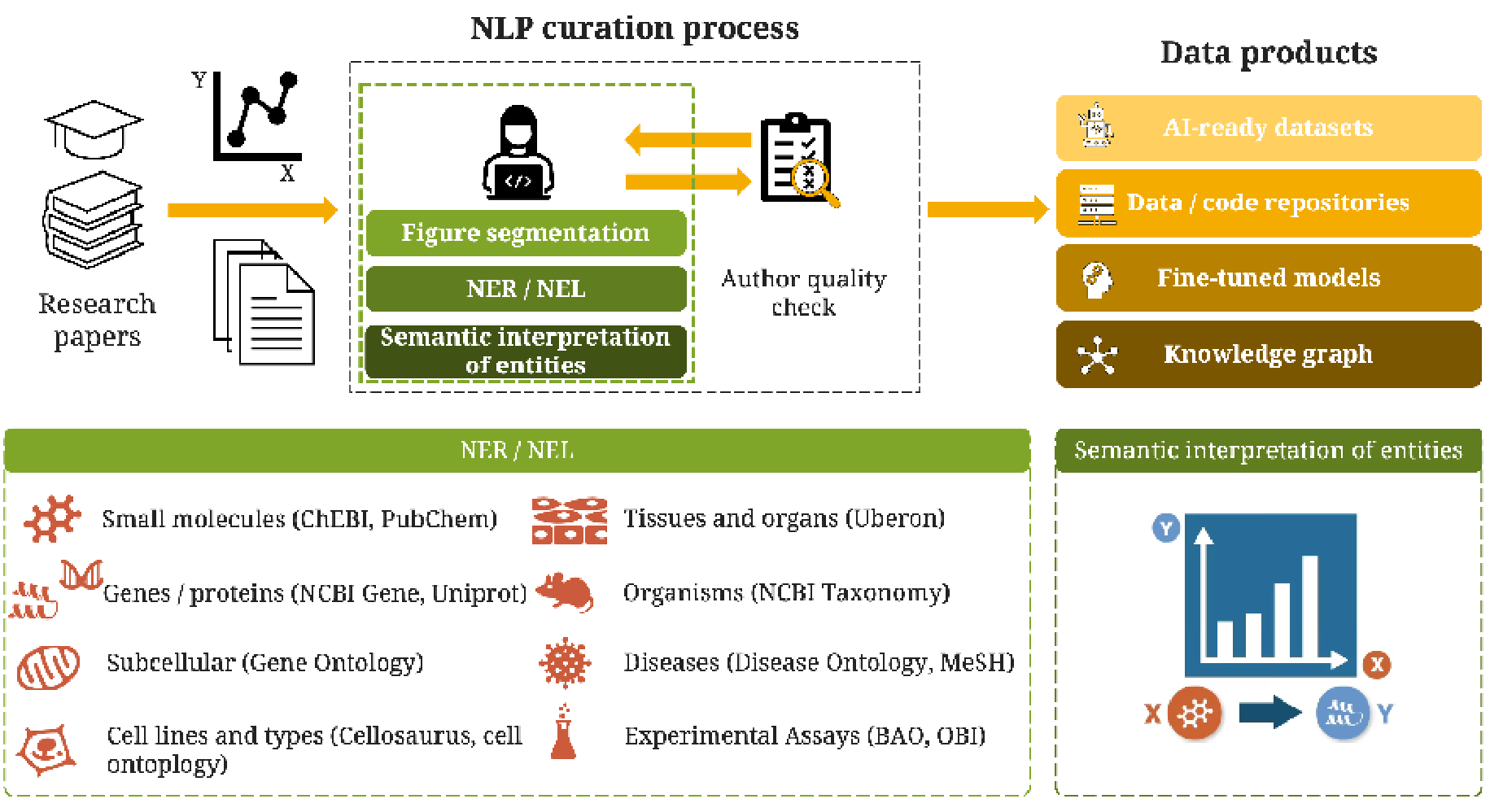}
    \caption{SourceData-NLP process for embedding article curation as part of the academic publishing process.}
    \label{fig:overall_process}
\end{figure}

The SourceData annotation is integrated into the publication workflow. It is therefore strongly constrained by the nature of the editorial process in a scientific journal. Consequently, the annotation framework is designed to balance simplicity and expressiveness, ensuring feasibility while reflecting key elements of the experimental approach \citep{sourcedata}. SourceData-NLP describes the data shown in scientific figures published in the field of cell and molecular biology. The annotations focus on biological entities relevant to the data's scientific meaning and the underlying experimental design. This approach streamlines the research process and makes detailed experimental data readily accessible. The annotation process, illustrated in Figure \ref{fig:overall_process}, includes the following steps:

\begin{enumerate}
    \item Splitting of composite figures into coherent panels (``figure segmentation'').
    \item Tagging and linking of biological entities to external identifiers (related to the NER and NEL tasks).
    \item Categorization of entities into the role they play in the specific experimental design.
\end{enumerate}

The workflow integrates a quality control feedback loop that leverages feedback from the original authors to ensure label accuracy. The annotation predominantly focuses on bioentities that are linked to identifiers from a selected number of ontologies (see Figure~\ref{fig:overall_process} and Online Material and Methods~\ref{app:online-methods}). To prioritize speed and efficiency, complex concepts such as biological processes, pathways, or detailed entity attributes are intentionally excluded. The details of the annotation process are further described in the Online Material and Methods~\ref{app:online-methods}, and annotation guidelines are provided in the Supplementary Material~\ref{app:guidelines}.

The SourceData-NLP dataset consists of 18,689 figures, manually divided into 62,543 annotated panels from 3,223 articles published across 25 journals. As the curation process is integrated into the ongoing publishing workflow at EMBO Press, the dataset is dominated by papers published in EMBO Journal, EMBO Reports, Molecular Systems Biology, and EMBO Molecular Medicine (2891 papers). The dataset is segmented into training, evaluation, and test sets in an 80-10-10 ratio. In total, 801,818 entities are provided, including 686,846 entities linked to identifiers in external databases. The distribution of these entities across the annotation classes can be found in Table~\ref{tab:mentions}. To our knowledge, SourceData-NLP represents one of the most extensive biomedical annotated datasets currently available for NER and NEL.

\begin{table}[]
    \small
    \centering
    \caption{Comprehensive Overview of Annotated Bioentities in the SourceData-NLP Dataset. This table presents a detailed breakdown of annotations, categorizing them by entity type for NER and the roles assigned within experimental contexts.
    Additionally, it provides statistics for NEL, including unique mentions tied to external identifiers. Percentages indicate the uniqueness of entity mentions within each category, reflecting the diversity of the dataset's contents.
    }
    \resizebox{\textwidth}{!}{%
    \begin{tabular}{l|rrr|rrr|rrr}
    \toprule
    \multicolumn{1}{c}{} & \multicolumn{3}{|c|}{\textbf{NER}}                                                         & \multicolumn{3}{c}{\textbf{Roles}}                                                                   & \multicolumn{3}{|c}{\textbf{NEL}}                                                         \\
    \multicolumn{1}{c}{} & \multicolumn{1}{|c}{} & \multicolumn{1}{c}{} & \multicolumn{1}{c|}{} & \multicolumn{1}{c}{Controlled}   & \multicolumn{1}{c}{Measured} & \multicolumn{1}{c}{} & \multicolumn{1}{|c}{} & \multicolumn{1}{c}{} & \multicolumn{1}{c}{} \\
    \multicolumn{1}{c}{} & \multicolumn{1}{|c}{Total}      & \multicolumn{1}{c}{Unique}       & \multicolumn{1}{c|}{\%}   & \multicolumn{1}{c}{Intervention} & \multicolumn{1}{c}{Assayed}  & \multicolumn{1}{c}{Other}      & \multicolumn{1}{|c}{Total}      & \multicolumn{1}{c}{Unique}       & \multicolumn{1}{c}{\%}   \\\midrule
    Small mol.           & 93\,582                   & 9\,089                     & 9.71                   & 45\,517                          & 15\,350                      & 32\,715                   & 87\,512                   & 4\,776                     & 5.46                   \\
    Gene products        & 355\,433                  & 29\,001                    & 8.16                   & 108\,577                         & 158\,048                     & 88\,808                   & 286\,561                  & 29\,244                    & 10.2                   \\
    Subcellular          & 44\,231                   & 3\,932                     & 8.89                   & 1\,243                           & 29\,218                      & 13\,770                   & 40\,109                   & 911                        & 2.27                   \\
    Cell type            & 31\,336                   & 2\,597                     & 8.28                   & 754                              & 13\,201                      & 17\,379                   & 31\,336                   & 558                        & 1.78                   \\
    Cell line            & 30\,457                   & 1\,724                     & 5.66                   & 266                              & 6\,647                       & 23\,544                   & 30\,441                   & 1\,064                     & 3.50                   \\
    Tissue               & 40\,677                   & 4\,356                     & 10.71                  & 578                              & 11\,959                      & 28\,140                   & 37\,381                   & 1\,437                     & 3.84                   \\
    Organism             & 48\,589                   & 2\,042                     & 4.20                   & 3\,278                           & 9\,082                       & 36\,229                   & 48\,050                   & 700                        & 1.46                   \\
    Disease              & 8\,045                    & 1\,280                     & 15.9                   &                                  &                              &                           & 7\,430                    & 567                        & 7.63                   \\
    Exp. Assay           & 149\,468                  & 13\,056                    & 8.73                   &                                  &                              &                           & 118\,026                  & 740                        & 0.63                   \\\midrule
    \textbf{Total}       & \textbf{801\,818}         & \textbf{67\,077}           & \textbf{8.37}          & \textbf{160\,213}                & \textbf{243\,505}            & \textbf{240\,585}         & \textbf{686\,846}         & \textbf{39\,997}           & \textbf{5.82}\\\bottomrule         
    \end{tabular}%
    }
    \label{tab:mentions}
        
\end{table}

\subsection{Finetuning language models for downstream NLP applications}

\subsubsection{NER}\label{sec:results-ner}

To illustrate the utility of the SourceData-NLP dataset, we evaluated the performance of two prominent biomedical pre-trained language models, PubMedBERT and BioLinkBERT, in the context of the NER task. We compared the base ($\sim110$ parameters) and large ($\sim330$ parameters) versions of these  models. The models were fine-tuned on the SourceData-NLP dataset using a multi-class training approach, where a single model assigns mutually exclusive labels across all entity categories. Details on the fine-tuning procedure can be found in the Supplementary Material~\ref{methods:ner-finetuning}.

The results of this experiment, summarized in Table~\ref{tab:ner}, demonstrate consistent incremental improvements with the larger versions of the models. For instance, the large model versions consistently outperform the base models across all entity categories, with notable enhancements in the large BioLinkBERT model compared to its base version in diseases (4.89\% improvement). The fact that the most significant improvement between the base and large methods corresponds to the, by far, least frequent entity type of SourceData-NLP highlights the generalization capacity of larger models. Comparison across entity classes reveals that the performance of the NER task varies significantly across different entity categories. The best performances are obtained with gene products, cell lines, organisms, and small molecules. Conversely, the performance is lowest for subcellular components, cell types, diseases, and experimental assays. There are several likely explanations for these differences. Entities that are easier to learn are also the most frequently annotated in our dataset. Additionally, entities such as gene products and cell lines tend to have simpler structures and morphologies (for example, `\verb 'Creb1'' or `\verb 'HeLa'' ) compared to the complex terminologies typical for subcellular structures, diseases, and cell types (for example, ``endoplasmic reticulum'' or ``hippocampal CA1 pyramidal neuron'').


\begin{table}
    \small
    \caption{Performance comparison of base and large model variants on the NER task. The results show the average F1 scores obtained through 5 inference rounds. The $\Delta$ column quantifies the performance gain of the large models over their base counterparts.}
    \label{tab:ner}
    \centering
    \begin{tabularx}{0.7\textwidth}{l|rrr|rrr|r}
        \toprule
        \textbf{Model} & \multicolumn{3}{c|}{\textbf{PubMedBERT}} & \multicolumn{3}{c|}{\textbf{BioLinkBERT}}    &        \\
        Model Size             & base     & large     & $\Delta (\%)$     & base     & large     & $\Delta (\%)$ & Support\\
        \midrule
        Gene product           & 91.9 & 92.3          & 0.43              & 89.2     & \textbf{92.7} & 3.78      & 26,321 \\
        Cell line              & 89.8 & 90.4          & 0.66              & 90.7     & \textbf{90.8} & 0.11      & 2,367  \\
        Organism               & 86.7 & \textbf{87.8} & 1.25              & 87.7     & 87.5          & -0.23     & 4,222  \\
        Small molecule         & 84.8 & 85.0          & 0.24              & 82.9     & \textbf{86.2} & 3.83      & 6,932  \\
        Tissue                 & 82.6 & \textbf{83.5} & 1.08              & 82.9     & 83.4          & 0.60      & 3,851  \\
        Subcellular            & 78.8 & 79.4          & 0.76              & \textbf{79.6} & 79.4     & -0.25     & 4,671  \\
        Cell type              & 72.1 & \textbf{72.6} & 0.69              & 71.4     & 72.2          & 1.11      & 2,872  \\
        Disease                & 66.6 & 68.6          & 2.92              & 66.1     & \textbf{69.5} & 4.89      & 602    \\
        Exp. Assay             & 67.3 & 67.8          & 0.74              & 67.1     & \textbf{68.7} & 2.33      & 11,196 \\
        \midrule
        \textbf{Micro avg.}    & 83.7 & 84.2          & 0.59              & 82.5     & \textbf{84.7} & 2.60      & 63,034 \\ 
        \textbf{Macro avg.}    & 80.1 & 80.8          & 0.87              & 79.7     & \textbf{81.2} & 1.85      & 63,034 \\
        \textbf{Weighted avg.} & 83.6 & 84.2          & 0.71              & 82.4     & \textbf{84.6} & 2.60      & 63,034  \\
        \bottomrule
    \end{tabularx}
\end{table}




The relative contribution of memorization versus generalization in NER transformer models is not fully understood. Previous work by \cite{Kim2021HowDY-memory} illustrates that models may not generalize well due to their propensity to take advantage of dataset biases or substandard naming conventions. In contrast, \cite{Tnzer2021MemorisationVG}  suggests that models can be robust to noisy data, assuming the patterns they learn are sufficiently frequent. To investigate these two aspects of the learning process, we evaluate the memorization and generalization capabilities of PubMedBERT and BioLinkBERT, both fine-tuned on the SourceData-NLP dataset. Following~\cite{Kim2021HowDY-memory}, we divided test set entities into ``memorization'' and ``generalization'' subsets. Entities were considered memorized if they were encountered during fine-tuning training or if they were present in the validation set. In contrast, the generalization subset is exclusively composed of entities that the model has never seen in either the training or the validation set.

\begin{figure}
    \centering
    \includegraphics[width=0.5\columnwidth]{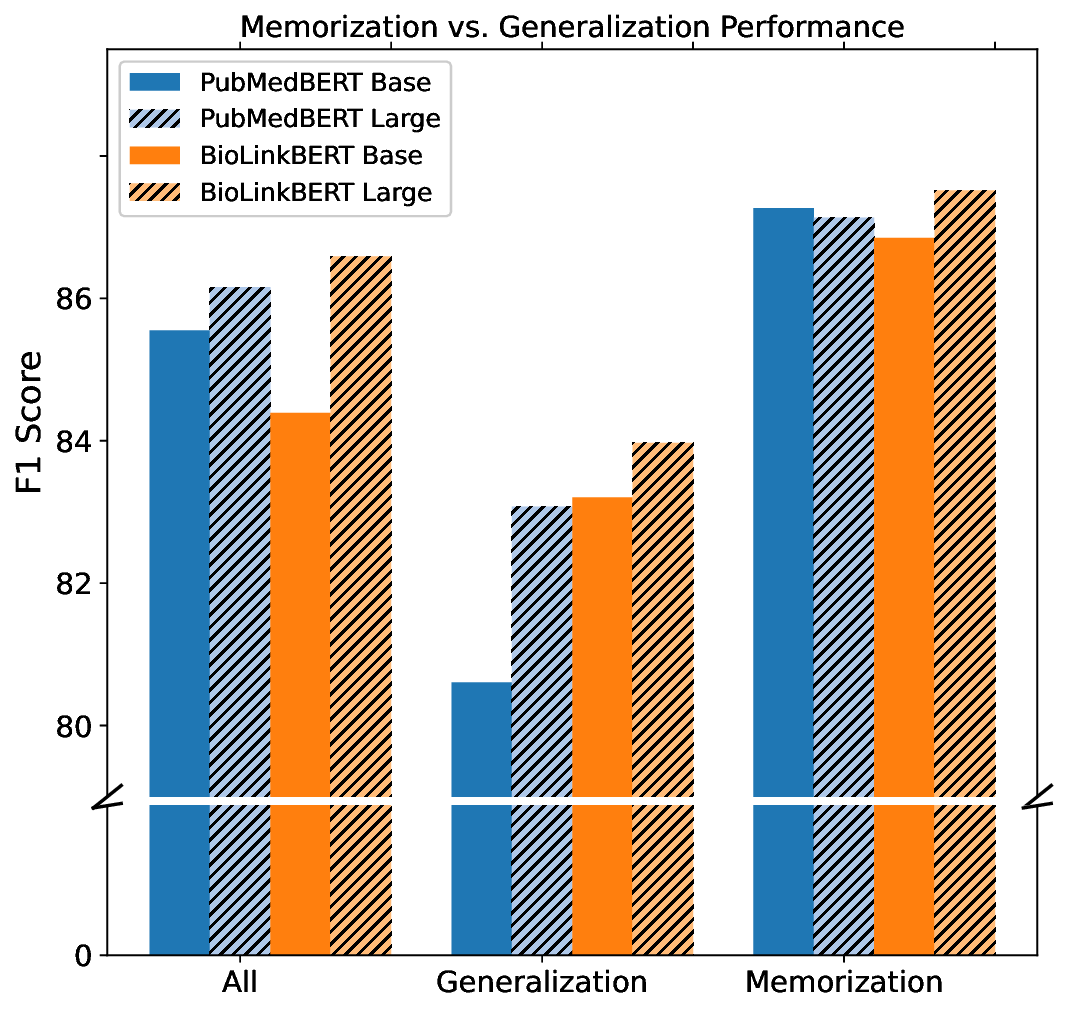}
    \caption{Memorization vs. generalization performance of PubMedBERT and BioLinkBERT models. The bar chart compares the F1 scores, distinguishing between overall performance, and specific performance in memorization and generalization tasks. Bars with diagonal stripes indicate the large versions of the models, whereas solid bars represent the base versions. PubMedBERT is denoted by blue bars, and BioLinkBERT by orange bars.
    }
    \label{fig:memo-vs-gen}
\end{figure}

The results are shown in Figure~\ref{fig:memo-vs-gen}. The findings reveal that PubMedBERT and BioLinkBERT achieve higher F1 scores on entities from the memorization subset than those from the generalization subset. Intriguingly, the base version of PubMedBERT shows slightly better memorization capabilities than its larger counterpart, which might suggest some degree of overfitting with the larger model. BioLinkBERT exhibits superior generalization capabilities with novel entities relative to PubMedBERT, indicating that its initial training regimen may have enhanced its ability to effectively handle unseen data.



\subsubsection{Semantic interpretation of empirical roles}\label{sec:results-roles}

In the context of the SourceData-NLP project, we introduced a novel NLP task to determine the empirical roles of entities within biomedical experiments. This token classification task assigns the ``measured variable'' or ``controlled variable'' roles to entities based on context. In this task, entities without one of these two roles are assigned the ``none'' category. This task is challenging, even for curators, as it requires interpreting the experimental design and is highly dependent on contextual information. We were therefore interested in investigating to which extent this task can be learned from the SourceData-NLP dataset.

Our experiments utilized both base and large versions of PubMedBERT. We primarily focused on gene products due to their balanced representation in the dataset (42\% as the ``controlled variable'' and 58\% as the ``measured variable'').

We explored three distinct approaches for training:
\begin{enumerate}
    \item\textbf{Context-only approach}: all gene products tagged in the SourceData-NLP dataset were first masked. The pre-trained models were then fine-tuned to classify these masked entities into ``measured variable'', ``controlled variable'' or ``none''. This approach guarantees that the models learn the classification task solely based on context, without the influence of the identity of the specific entities.
    \item \textbf{Marked-entity approach}: gene products were first flanked by a special token to mark their position in the text explicitly. The pre-trained models were then fine-tuned \textit{without} any entity masking. This allows the models to learn both from context and from the identity of the entities.
    \item \textbf{Single-step approach}: in contrast to the previous approaches, here entities were not masked and their position was not marked. The models were therefore trained to both recognize the entity and classify their role in a single step, thus combining NER with semantic role classification in a joint task.
\end{enumerate}

\begin{table}[t!]
    \caption{Micro averaged F1 scores of the empirical role of entities task for gene products for the context only, marked-entity, and single-step approaches.}
    \label{tab:results-roles}
    \begin{tabularx}{\textwidth}{clllll}
        \toprule
                                   & \multicolumn{5}{c}{Micro averaged F1}                                                 \\
        \cline{2-6} 
        \textbf{Model}             & Context-only & Context only corr. & Marked-entity & Marked-entity corr. & Single-step \\
        \midrule
        \textbf{PubMedBERT-base}   & 91.2         & 83.8               & 91.6          & 84.2                & 81.4        \\
        \textbf{PubMedBERT-large}  & 91.0         & 84.0               & 92.3          & 85.2                & 82.5        \\
        \bottomrule
    \end{tabularx}
\end{table}

In the inference stage, a prior NER step is required to identify entity mentions in the text for both context-only and marked-entity tasks. However, the single-step approach handles both tasks simultaneously, eliminating the need for a multi-stage pipeline workflow.

The results of these experiments, detailed in Table~\ref{tab:results-roles}, validated the efficacy of our novel approaches. Particularly, the models employing the marked-entity and context-only approaches demonstrated robust capabilities in accurately identifying the nuanced roles of bioentities within complex experimental setups. Since these first two approaches require a prior NER step during inference, we also report a combined score for the pipeline by multiplying the F1 scores of the role classification step by the respective NER scores, as detailed in Table~\ref{tab:ner}. The corrected F1 scores across different approaches ranged from 81.4\% to 84.2\%, indicating the potential for high accuracy in the contextual interpretation of biomedical entities. This highlights the unique capability of SourceData-NLP for accurately representing causal hypothesis testing in literature.

\subsubsection{Multimodal segmentation of compound figures}\label{sec:panelization}

We used the SourceData-NLP dataset to generate a dataset to build a multimodal pipeline that separates compound scientific figures into their constituent panels and matches them to the corresponding panel captions. We followed a two-step procedure to achieve this goal. First, we used object detection algorithms to separate the figure into its panels. Second, we used a multimodal LLM to extract the correspondent panel description from the figure caption, ensuring that the panel caption is understandable on its own without the need for the context of the full figure caption.

We generated a dataset containing 13039 figures with box annotated panels (divided on 11735 training, 651 validation, and 653 test examples), also annotated with the different panels. We use the boxes of the panels to fine-tune the pre-trained YOLOv10~\citep{wang2024yolov10} object detection model. We evaluated the model using the metrics mAP$_{50}$ and mAP$_{50-95}$. The former represents the mean average precision at the intersection-over-union (IoU) threshold of 0.5, meaning that if a prediction overlaps 50\% with the ground truth, it is considered a success. The mAP$_{50-95}$ metric provides a more comprehensive evaluation of the model's performance, averaging the precision across various levels of overlap (from 50\% to 95\% at steps of 5) between predictions and ground truth. The fine-tuned model achieves a mAP$_{50} = 98.2\%$ and mAP$_{50-95} = 87.0\%$ in the test dataset. 

To map each segmented panel to their corresponding part of the caption, we used API calls to the OpenAI GPT-4o multimodal model. Each API call contained the text of a full figure caption and the image of one of the panels of the correspondent figure extracted with our image segmentation model described above. The results returned by the model are the extracted panel description minimally edited to be fully understood without the need for context from the full figure caption. The system prompt used for the LLM is shown in appendix~\ref{app:segmentation}. The accuracy of this second step for caption-panel matching is 97.4\%. Here we count as a correct result every panel assigned the correct panel label, while every panel with the false label assignment is considered a miss. The accuracy is measured only on those panels correctly identified by the object detection algorithm.

Prior work using ImageCLEF2016 has reported accuracies of up to mAP$_{50} = 90\%$ and mAP$_{50-95} = 72\%$~\citep{Jiang2021ATF}, achieved by finetuning YOLOv3~\citep{Adarsh2020YOLOVO}. However, direct comparison with these works is not possible because they were based on the medical task of ImageCLEF2016~\citep{Herrera2016OverviewOT}, which focuses on extracting images from scientific figures while we are extracting the entire panel, including the image and the labels.

\section{Discussion}

The SourceData-NLP shows the benefits of integrating curation within the publishing process. Through our  figure-centric annotation approach, we have produced one of the most extensive NER and NEL datasets currently available in the field of molecular biology and life sciences. The scale and utility of SourceData-NLP for training large language models emphasizes the potential of this approach to  enhance the accessibility and reusability of scientific data but also accelerates the development of AI tools for scientific discovery.

By focusing the annotations on figures, as opposed to processing papers in their full length, the curation effort was manageable and could be sustained over an extended period. Moreover, by concentrating on figures, we created a unique multimodal dataset that links the images of individual experimental results shown in figure panels to the corresponding segments of their captions. This innovative structure enables exploring novel AI tasks, such as figure-caption matching, image-based information retrieval, and multimodal vector representations.

The SourceData-NLP contains more than 600,000 disambiguated entities present in 62,543 annotated figure panels from 3,223 papers, making it a useful resource for training NLP models. Its diverse representation of biomedical entities ensures that resultant models are adept at addressing challenges inherent to the field. To facilitate researchers' access to SourceData-NLP and assist the advancement of biomedical NER and NEL methods, we provide the source code for the raw data, machine-learning-ready data, and models generated. 

Additionally, we have delineated a novel NLP task centered on the interpretation of the role of entities in a given experimental design. We show that this task, which depends heavily on contextual information, can be efficiently learned to determine whether a gene product is measured (a "measured variable") or whether it is the target of an experimental perturbation (a "controlled variable"). Identifying these roles within a specific experiment allows us to infer the causal hypothesis being tested (for example, "Does the perturbation of gene product X influence the measurement of gene product Y?"). The discrimination of controlled and experimental variables is a unique property of SourceData-NLP that greatly enhances its utility.

In biomedical research, testing causal hypotheses is a key experimental approach to investigating the molecular mechanisms underlying biological processes and human diseases. The models trained on SourceData-NLP thus allow for the large-scale extraction of such causal hypotheses from the literature and the traceability of their links to published scientific results.

Over the years, significant efforts have been made to create annotated datasets for training and benchmarking algorithms on NER and NEL tasks~\citep[see bigBio][for a detailed summary of these efforts]{bigbio}. The recent BioRED dataset~\citep{biored} has a similar scope to SourceData-NLP. It annotates genes, diseases, chemicals, genetic variants, species, and cell lines. However, BioRED is focused on the relation detection (RD) task, which reveals interactions between entities. BioRED has a set of more than 20,000 entities tagged with links to databases and includes a novelty classification label intended to highlight novel discoveries. Another recently published dataset is the Europe PMC Annotated Full-text Corpus~\citep{Yang2023EuropePA}. They report a total of over 72,000 annotated entities of the classes gene/protein, disease, and organism. Their data sample covers isolated sentences from the 300 selected full-text articles. The annotations where done using the triple-anonymous annotation procedure, with no involvement of the authors of the manuscripts. With these properties, the scope of he Europe PMC Annotated Full-text Corpus differs from that of the SourceData-NLP dataset.

Previous studies typically focus on accurately extracting entities (NER/NEL) and their relationships by annotating sentences, abstracts, or full texts. The unique advantage of SourceData-NLP lies in its innovative approach to annotating figure captions that describe scientific results and in defining the roles that biomedical entities play within specific experimental designs. Moreover, SourceData-NLP encompasses a wider set of entity classes, offering more comprehensive coverage that enhances the dataset’s utility across various biomedical research domains. Finally, the SourceData-NLP dataset is intrinsically multimodal since the annotated text is paired with its corresponding scientific figures and panels.

To the best of our knowledge, previous work has not extensively explored such detailed annotation of empirical roles of bioentities in scientific literature, along with the expansive array of tagged entity classes and a multimodal format. This novel aspect of SourceData-NLP facilitates the creation of data models that capture critical features of scientific experiments in biomedical research, thereby providing a deeper and more functionally relevant layer of data that enhances the potential for breakthrough discoveries in the field.

\subsection{Limitations and future work}\label{sec:future} 

One limitation of the SourceData-NLP dataset is the inherent noise of human-based curation. The constraints of the publishing process did not allow curation by multiple annotators. Even though an independent validation step by authors and quality control by the editorial office were in place, all discrepancies could not be eliminated. While the SourceData-NLP is already a robust resource, we anticipate that the workflow will benefit from enhanced automated consistency checks.

The scope of the journals from which the dataset was generated is focused on cell and molecular biology. While it provides a rich resource for entities such as small molecules, genes, proteins, subcellular components, and cell lines, its coverage of other biomedical entities like cell types, tissues, organs, and diseases is sparser. A promising avenue to overcome such limitations is to use data augmentation strategies when training models. One possibility is using generative models by generating synthetic examples from an exhaustive entity dictionary. These synthetic examples could considerably expand the dataset to improve the coverage and robustness of fine-tuned models~\citep[e.g., ][]{guo2023improving,whitehouse2023llmpowered,yuan2023large}. An alternative strategy is to use existing examples as templates in which a fraction of the annotated entities could be replaced by sampling terms of the appropriate type from dictionaries or controlled vocabularies in a process akin to ontology-guide data augmentation \citep{Abdollahi2020OntologyGuidedDA}. This approach would provide a richer and more varied set of examples, potentially enhancing the dataset's depth. A third approach to this problem could be to merge SourceData-NLP with other NER datasets dedicated to the different classes annotated in SourceData-NLP and train the models using partial annotation training~\citep{Ding2023PartialAL}.

While treating the tagging and characterization of biomedical entities as a sequential pipeline shows promising results, it is affected by error propagation. The rise of large language models (LLMs) such as GPT and others, has demonstrated their capabilities in various NLP tasks when formulated as text-to-text tasks \citep{t5}. However, their performance on NER tasks has been reported as still remaining subpar~\citep{jimenez22}, especially in the biomedical field~\citep{deusser2023informed}. The rapid progress in LLMs suggests, however, that the text-to-text approach with generative models may improve in the future. 

An exciting potential application of SourceData-NLP is the construction of a large knowledge graph tailored for molecular biology. Knowledge graphs are structured representations of information where entities and their interrelationships are depicted as nodes and edges, respectively. With the semantic roles obtained from our novel task, we can systematically identify the experimental roles of biological entities in scientific results published in the field of molecular biology literature. This enables the representation of the experimental design in terms of entities linked to the respective "controlled variable" and "measured variable". It is therefore possible to build a knowledge graph with entities as nodes and directional relationships that represent that causal hypothesis tested in the reported experiments. Such a knowledge graph would serve as a visual and interactive tool for researchers and facilitate graph data mining, hypothesis generation, and predictive modeling.

\section*{Materials and methods}
Following the manuscript submission guidelines of Nature Methods, we included the materials and methods section as an online supplementary material. Please see Supplementary Material~\ref{app:online-methods}.

\section*{Acknowledgements}

We thank David Kartchner from the Georgia Institute of Technology for his help with the dataset deposition to the BigBIO biomedical dataset library. We thank Robin Liechti and Lou Götz from the Swiss Institute of Bioinformatics for their help with the SourceData core database, and Alejandro Riera from EMBO for his support with the GPU computing infrastructure. The authors used AI writing assistants (GPT4, Claude3, Grammarly) to draft and edit the text of the manuscript; the entire text was proofread by the authors.

\section*{Data Availability Section}
The data and models underlying this study are available in the following resources:

\begin{itemize}
    \item Scripts to generate the SourceData-NLP from the raw SourceData annotations: https://github.com/source-data/soda-data
    \item SourceData-NLP in pre-tokenized format : https://huggingface.co/datasets/EMBO/SourceData 
    \item Fine-tuned NLP models: https://github.com/source-data/soda-model
    \item Fine-tuned object detection model: https://github.com/source-data/soda\_image\_segmentation
\end{itemize}

\bibliographystyle{natbib}
\bibliography{main}

\begin{thebibliography}{}

\bibitem[Abdollahi {\em et~al.}(2020)Abdollahi, Gao, Mei, Ghosh, and Li]{Abdollahi2020OntologyGuidedDA}
Abdollahi, M.  {\em et~al.} (2020).
\newblock Ontology-guided data augmentation for medical document classification.
\newblock In {\em Conference on Artificial Intelligence in Medicine in Europe\/}.

\bibitem[Adarsh {\em et~al.}(2020)Adarsh, Rathi, and Kumar]{Adarsh2020YOLOVO}
Adarsh, P.  {\em et~al.} (2020).
\newblock Yolo v3-tiny: Object detection and recognition using one stage improved model.
\newblock {\em 2020 6th International Conference on Advanced Computing and Communication Systems (ICACCS)\/}, pages 687--694.

\bibitem[Ashburner {\em et~al.}(2000)Ashburner, Ball, Blake, Botstein, Butler, Cherry, Davis, Dolinski, Dwight, Eppig, Harris, Hill, Issel-Tarver, Kasarskis, Lewis, Matese, Richardson, Ringwald, Rubin, and Sherlock]{geneont1}
Ashburner, M.  {\em et~al.} (2000).
\newblock Gene ontology: tool for the unification of biology.
\newblock {\em Nature Genetics\/}, {\bf 25}, 25--29.

\bibitem[Bairoch(2018)Bairoch]{cellosaurus}
Bairoch, A. (2018).
\newblock The cellosaurus, a cell-line knowledge resource.
\newblock {\em Journal of biomolecular techniques : JBT\/}, {\bf 29 2}, 25--38.

\bibitem[Bandrowski {\em et~al.}(2016)Bandrowski, Brinkman, Brochhausen, Brush, Bug, Chibucos, Clancy, Courtot, Derom, Dumontier, Fan, Fostel, Fragoso, Gibson, Gonzalez-Beltran, Haendel, He, Heiskanen, Hernandez-Boussard, Jensen, Lin, Lister, Lord, Malone, Manduchi, McGee, Morrison, Overton, Parkinson, Peters, Rocca-Serra, Ruttenberg, Sansone, Scheuermann, Schober, Smith, Soldatova, Stoeckert, Taylor, Torniai, Turner, Vita, Whetzel, and Zheng]{obi}
Bandrowski, A.  {\em et~al.} (2016).
\newblock The ontology for biomedical investigations.
\newblock {\em PLOS ONE\/}, {\bf 11}(4), 1--19.

\bibitem[Bethesda(2004)Bethesda]{ncbigene}
Bethesda (2004).
\newblock {\em NCBI Gene\/}.
\newblock National Library of Medicine (US).

\bibitem[Bethesda(2005)Bethesda]{pubmedhelp}
Bethesda (2005).
\newblock {\em PubMed Help\/}.
\newblock National Center for Biotechnology Information (US).

\bibitem[Bikel {\em et~al.}(1997)Bikel, Miller, Schwartz, and Weischedel]{bikel-etal-1997-nymble}
Bikel, D.~M.  {\em et~al.} (1997).
\newblock {N}ymble: a high-performance learning name-finder.
\newblock In {\em Fifth Conference on Applied Natural Language Processing\/}, pages 194--201, Washington, DC, USA. Association for Computational Linguistics.

\bibitem[Carbon {\em et~al.}(2021)Carbon, Douglass, Good, Unni, Harris, Mungall, Basu, Chisholm, Dodson, Hartline, Fey, Thomas, Albou, Ebert, Kesling, Mi, Muruganujan, Huang, Mushayahama, LaBonte, Siegele, Antonazzo, Attrill, Brown, Garapati, Marygold, Trovisco, dos Santos, Falls, Tabone, Zhou, Goodman, Strelets, Thurmond, Garmiri, Ishtiaq, Rodr{\'i}guez-L{\'o}pez, Acencio, Kuiper, L{\ae}greid, Logie, Lovering, Kramarz, Saverimuttu, Pinheiro, Gunn, Su, Thurlow, Chibucos, Giglio, Nadendla, Munro, Jackson, Duesbury, del Toro, Meldal, Paneerselvam, Perfetto, Porras, Orchard, Shrivastava, Chang, Finn, Mitchell, Rawlings, Richardson, Sangrador-Vegas, Blake, Christie, Dolan, Drabkin, Hill, Ni, Sitnikov, Harris, Oliver, Rutherford, Wood, Hayles, B{\"a}hler, Bolton, DePons, Dwinell, Hayman, Kaldunski, Kwitek, Laulederkind, Plasterer, Tutaj, Vedi, Wang, D’Eustachio, Matthews, Balhoff, Aleksander, Alexander, Cherry, Engel, Gondwe, Karra, Miyasato, Nash, Simison, Skrzypek, Weng, Wong, Feuermann, Gaudet, Morgat,
  Bakker, Berardini, Reiser, Subramaniam, Huala, Arighi, Auchincloss, Axelsen, Argoud-Puy, Bateman, Blatter, Boutet, Bowler, Breuza, Bridge, Britto, Bye-A-Jee, Casals-Casas, Coudert, Denny, Estreicher, Famiglietti, Georghiou, Gos, Gruaz-Gumowski, Hatton-Ellis, Hulo, Ignatchenko, Jungo, Laiho, Mercier, Lieberherr, Lock, Lussi, MacDougall, Magrane, Martin, Masson, Natale, Hyka-Nouspikel, Pedruzzi, Pourcel, Poux, Pundir, Rivoire, Speretta, Sundaram, Tyagi, Warner, Zaru, Wu, Diehl, Chan, Grove, Lee, M{\"u}ller, Raciti, Auken, Sternberg, Berriman, Paulini, Howe, Gao, Wright, Stein, Howe, Toro, Westerfield, Jaiswal, Cooper, and Elser]{genont2}
Carbon, S.  {\em et~al.} (2021).
\newblock The gene ontology resource: enriching a gold mine.
\newblock {\em Nucleic Acids Research\/}, {\bf 49}, D325 -- D334.

\bibitem[Consortium(2019)Consortium]{uniprot}
Consortium, T.~U. (2019).
\newblock Uniprot: a worldwide hub of protein knowledge.
\newblock {\em Nucleic Acids Research\/}, {\bf 47}, D506 -- D515.

\bibitem[de~Herrera {\em et~al.}(2016)de~Herrera, Schaer, Bromuri, and M{\"u}ller]{Herrera2016OverviewOT}
de~Herrera, A. G.~S.  {\em et~al.} (2016).
\newblock Overview of the imageclef 2016 medical task.
\newblock In {\em Conference and Labs of the Evaluation Forum\/}.

\bibitem[Deußer {\em et~al.}(2023)Deußer, Hillebrand, Bauckhage, and Sifa]{deusser2023informed}
Deußer, T.  {\em et~al.} (2023).
\newblock Informed named entity recognition decoding for generative language models.

\bibitem[Devlin {\em et~al.}(2018)Devlin, Chang, Lee, and Toutanova]{bert}
Devlin, J.  {\em et~al.} (2018).
\newblock Bert: Pre-training of deep bidirectional transformers for language understanding.

\bibitem[Diehl {\em et~al.}(2016)Diehl, Meehan, Bradford, Brush, Dahdul, Dougall, He, Osumi-Sutherland, Ruttenberg, Sarntivijai, Slyke, Vasilevsky, Haendel, Blake, and Mungall]{cellont}
Diehl, A.~D.  {\em et~al.} (2016).
\newblock The cell ontology 2016: enhanced content, modularization, and ontology interoperability.
\newblock {\em Journal of Biomedical Semantics\/}, {\bf 7}.

\bibitem[Ding {\em et~al.}(2023)Ding, Colavizza, and Zhang]{Ding2023PartialAL}
Ding, L.  {\em et~al.} (2023).
\newblock Partial annotation learning for biomedical entity recognition.
\newblock {\em ArXiv\/}, {\bf abs/2305.13120}.

\bibitem[Fries {\em et~al.}(2022)Fries, Weber, Seelam, Altay, Datta, Garda, Kang, Su, Kusa, Cahyawijaya, {\em et~al.}]{bigbio}
Fries, J.~A.  {\em et~al.} (2022).
\newblock Bigbio: A framework for data-centric biomedical natural language processing.
\newblock {\em arXiv preprint arXiv:2206.15076\/}.

\bibitem[Griffiths-Jones {\em et~al.}(2003)Griffiths-Jones, Bateman, Marshall, Khanna, and Eddy]{rfam}
Griffiths-Jones, S.  {\em et~al.} (2003).
\newblock Rfam: an rna family database.
\newblock {\em Nucleic acids research\/}, {\bf 31 1}, 439--41.

\bibitem[Gu {\em et~al.}(2022)Gu, Tinn, Cheng, Lucas, Usuyama, Liu, Naumann, Gao, and Poon]{pubmedbert}
Gu, Y.  {\em et~al.} (2022).
\newblock Domain-specific language model pretraining for biomedical natural language processing.
\newblock {\em {ACM} Transactions on Computing for Healthcare\/}, {\bf 3}(1), 1--23.

\bibitem[Guo {\em et~al.}(2023)Guo, Wang, Wang, and Yu]{guo2023improving}
Guo, Z.  {\em et~al.} (2023).
\newblock Improving small language models on pubmedqa via generative data augmentation.

\bibitem[Gururangan {\em et~al.}(2020)Gururangan, Marasović, Swayamdipta, Lo, Beltagy, Downey, and Smith]{biomedroberta}
Gururangan, S.  {\em et~al.} (2020).
\newblock Don't stop pretraining: Adapt language models to domains and tasks.
\newblock In {\em Proceedings of ACL\/}.

\bibitem[Hastings {\em et~al.}(2015)Hastings, Owen, Dekker, Ennis, Kale, Muthukrishnan, Turner, Swainston, Mendes, and Steinbeck]{chebi}
Hastings, J.  {\em et~al.} (2015).
\newblock {ChEBI in 2016: Improved services and an expanding collection of metabolites}.
\newblock {\em Nucleic Acids Research\/}, {\bf 44}(D1), D1214--D1219.

\bibitem[Hoffart {\em et~al.}(2011)Hoffart, Yosef, Bordino, F{\"u}rstenau, Pinkal, Spaniol, Taneva, Thater, and Weikum]{hoffart-etal-2011-robust}
Hoffart, J.  {\em et~al.} (2011).
\newblock Robust disambiguation of named entities in text.
\newblock In {\em Proceedings of the 2011 Conference on Empirical Methods in Natural Language Processing\/}, pages 782--792, Edinburgh, Scotland, UK. Association for Computational Linguistics.

\bibitem[Jiang {\em et~al.}(2021)Jiang, Schwenker, Spreadbury, Ferrier, Chan, and Cossairt]{Jiang2021ATF}
Jiang, W.  {\em et~al.} (2021).
\newblock A two-stage framework for compound figure separation.
\newblock {\em 2021 IEEE International Conference on Image Processing (ICIP)\/}, pages 1204--1208.

\bibitem[{Jim{\'e}nez Guti{\'e}rrez} {\em et~al.}(2022){Jim{\'e}nez Guti{\'e}rrez}, {McNeal}, {Washington}, {Chen}, {Li}, {Sun}, and {Su}]{jimenez22}
{Jim{\'e}nez Guti{\'e}rrez}, B.  {\em et~al.} (2022).
\newblock {Thinking about GPT-3 In-Context Learning for Biomedical IE? Think Again}.
\newblock {\em arXiv e-prints\/}, page arXiv:2203.08410.

\bibitem[Kim and Kang(2021)Kim and Kang]{Kim2021HowDY-memory}
Kim, H. and Kang, J. (2021).
\newblock How do your biomedical named entity recognition models generalize to novel entities?
\newblock {\em IEEE Access\/}, {\bf 10}, 31513--31523.

\bibitem[Kim {\em et~al.}(2021)Kim, Chen, Cheng, Gindulyte, He, He, Li, Shoemaker, Thiessen, Yu, Zaslavsky, Zhang, and Bolton]{pubchem}
Kim, S.  {\em et~al.} (2021).
\newblock Pubchem in 2021: new data content and improved web interfaces.
\newblock {\em Nucleic Acids Research\/}, {\bf 49}, D1388 -- D1395.

\bibitem[Lee {\em et~al.}(2019)Lee, Yoon, Kim, Kim, Kim, So, and Kang]{biobert}
Lee, J.  {\em et~al.} (2019).
\newblock {BioBERT}: a pre-trained biomedical language representation model for biomedical text mining.
\newblock {\em Bioinformatics\/}.

\bibitem[Li {\em et~al.}(2021)Li, Jiang, Zhang, Trabucco, Raciti, Smith, Ringwald, Marai, Arighi, and Shatkay]{Li2021UtilizingIA}
Li, P.  {\em et~al.} (2021).
\newblock Utilizing image and caption information for biomedical document classification.
\newblock {\em Bioinformatics\/}, {\bf 37}, i468 -- i476.

\bibitem[Liechti {\em et~al.}(2016)Liechti, George, El-Gebali, G{\"o}tz, Crespo, Xenarios, and Lemberger]{sourcedata}
Liechti, R.  {\em et~al.} (2016).
\newblock Sourcedata - a semantic platform for curating and searching figures.
\newblock {\em bioRxiv\/}.

\bibitem[Lipscomb(2000)Lipscomb]{mesh}
Lipscomb, C.~E. (2000).
\newblock Medical subject headings (mesh).
\newblock {\em Bulletin of the Medical Library Association\/}, {\bf 88 3}, 265--6.

\bibitem[Liu {\em et~al.}(2019)Liu, Ott, Goyal, Du, Joshi, Chen, Levy, Lewis, Zettlemoyer, and Stoyanov]{roberta}
Liu, Y.  {\em et~al.} (2019).
\newblock Roberta: {A} robustly optimized {BERT} pretraining approach.
\newblock {\em CoRR\/}, {\bf abs/1907.11692}.

\bibitem[Luo {\em et~al.}(2022)Luo, Lai, Wei, Arighi, and Lu]{biored}
Luo, L.  {\em et~al.} (2022).
\newblock {BioRED: a rich biomedical relation extraction dataset}.
\newblock {\em Briefings in Bioinformatics\/}, {\bf 23}(5).
\newblock bbac282.

\bibitem[Mungall {\em et~al.}(2011)Mungall, Torniai, Gkoutos, Lewis, and Haendel]{uberon}
Mungall, C.~J.  {\em et~al.} (2011).
\newblock Uberon, an integrative multi-species anatomy ontology.
\newblock {\em Genome Biology\/}, {\bf 13}, R5 -- R5.

\bibitem[Perera {\em et~al.}(2020)Perera, Dehmer, and Emmert-Streib]{perera}
Perera, N.  {\em et~al.} (2020).
\newblock Named entity recognition and relation detection for biomedical information extraction.
\newblock {\em Frontiers in Cell and Developmental Biology\/}, {\bf 8}.

\bibitem[Raffel {\em et~al.}(2020)Raffel, Shazeer, Roberts, Lee, Narang, Matena, Zhou, Li, and Liu]{t5}
Raffel, C.  {\em et~al.} (2020).
\newblock Exploring the limits of transfer learning with a unified text-to-text transformer.
\newblock {\em Journal of Machine Learning Research\/}, {\bf 21}(140), 1--67.

\bibitem[Ramshaw and Marcus(1995)Ramshaw and Marcus]{biotagging}
Ramshaw, L.~A. and Marcus, M.~P. (1995).
\newblock Text chunking using transformation-based learning.
\newblock {\em ArXiv\/}, {\bf cmp-lg/9505040}.

\bibitem[Sayers {\em et~al.}(2021)Sayers, Cavanaugh, Clark, Pruitt, Schoch, Sherry, and Karsch-Mizrachi]{ncbitaxonomy2}
Sayers, E.~W.  {\em et~al.} (2021).
\newblock Genbank.
\newblock {\em Nucleic Acids Research\/}, {\bf 42}, D32 -- D37.

\bibitem[Schoch {\em et~al.}(2020)Schoch, Ciufo, Domrachev, Hotton, Kannan, Khovanskaya, Leipe, McVeigh, O'Neill, Robbertse, Sharma, Soussov, Sullivan, Sun, Turner, and Karsch-Mizrachi]{ncbitaxonomy1}
Schoch, C.~L.  {\em et~al.} (2020).
\newblock Ncbi taxonomy: a comprehensive update on curation, resources and tools.
\newblock {\em Database : the journal of biological databases and curation\/}, {\bf 2020}.

\bibitem[Schriml {\em et~al.}(2011)Schriml, Arze, Nadendla, Chang, Mazaitis, Felix, Feng, and Kibbe]{do}
Schriml, L.~M.  {\em et~al.} (2011).
\newblock Disease ontology: a backbone for disease semantic integration.
\newblock {\em Nucleic Acids Research\/}, {\bf 40}, D940 -- D946.

\bibitem[Sever {\em et~al.}(2019)Sever, Roeder, Hindle, Sussman, Black, Argentine, Manos, and Inglis]{biorxiv}
Sever, R.  {\em et~al.} (2019).
\newblock biorxiv: the preprint server for biology.
\newblock {\em bioRxiv\/}.

\bibitem[Shin {\em et~al.}(2020)Shin, Zhang, Bakhturina, Puri, Patwary, Shoeybi, and Mani]{biomegatron}
Shin, H.-C.  {\em et~al.} (2020).
\newblock Biomegatron: Larger biomedical domain language model.

\bibitem[T{\"a}nzer {\em et~al.}(2021)T{\"a}nzer, Ruder, and Rei]{Tnzer2021MemorisationVG}
T{\"a}nzer, M.  {\em et~al.} (2021).
\newblock Memorisation versus generalisation in pre-trained language models.
\newblock In {\em Annual Meeting of the Association for Computational Linguistics\/}.

\bibitem[Vaswani {\em et~al.}(2017)Vaswani, Shazeer, Parmar, Uszkoreit, Jones, Gomez, Kaiser, and Polosukhin]{transformers}
Vaswani, A.  {\em et~al.} (2017).
\newblock Attention is all you need.

\bibitem[Visser {\em et~al.}(2011)Visser, Abeyruwan, Vempati, Smith, Lemmon, and Sch{\"u}rer]{bao}
Visser, U.  {\em et~al.} (2011).
\newblock Bioassay ontology (bao): a semantic description of bioassays and high-throughput screening results.
\newblock {\em BMC Bioinformatics\/}, {\bf 12}, 257 -- 257.

\bibitem[Wang {\em et~al.}(2024)Wang, Chen, Liu, Chen, Lin, Han, and Ding]{wang2024yolov10}
Wang, A.  {\em et~al.} (2024).
\newblock Yolov10: Real-time end-to-end object detection.
\newblock {\em arXiv preprint arXiv:2405.14458\/}.

\bibitem[Whitehouse {\em et~al.}(2023)Whitehouse, Choudhury, and Aji]{whitehouse2023llmpowered}
Whitehouse, C.  {\em et~al.} (2023).
\newblock Llm-powered data augmentation for enhanced crosslingual performance.

\bibitem[Yang {\em et~al.}(2023)Yang, Saha, Venkatesan, Tirunagari, Vartak, and McEntyre]{Yang2023EuropePA}
Yang, X.  {\em et~al.} (2023).
\newblock Europe pmc annotated full-text corpus for gene/proteins, diseases and organisms.
\newblock {\em Scientific Data\/}, {\bf 10}.

\bibitem[Yasunaga {\em et~al.}(2022)Yasunaga, Leskovec, and Liang]{biolinkbert}
Yasunaga, M.  {\em et~al.} (2022).
\newblock Linkbert: Pretraining language models with document links.
\newblock In {\em Association for Computational Linguistics (ACL)\/}.

\bibitem[Yuan {\em et~al.}(2023)Yuan, Tang, Jiang, and Hu]{yuan2023large}
Yuan, J.  {\em et~al.} (2023).
\newblock Large language models for healthcare data augmentation: An example on patient-trial matching.

\bibitem[Zhao {\em et~al.}(2020)Zhao, Su, Lu, and Wang]{zhao20-biomedical-curation}
Zhao, S.  {\em et~al.} (2020).
\newblock {Recent advances in biomedical literature mining}.
\newblock {\em Briefings in Bioinformatics\/}, {\bf 22}(3).
\newblock bbaa057.

\end{thebibliography}

\newpage
\appendix
\renewcommand{\thesection}{\Alph{section}}

\makeatletter
\renewcommand{\fnum@figure}{Supplementary Figure \thefigure}
\renewcommand{\fnum@table}{Supplementary Table \thetable}
\makeatother

\setcounter{figure}{1}
\renewcommand{\thefigure}{\arabic{figure}}

\setcounter{table}{1}
\renewcommand{\thetable}{\arabic{figure}}

\section{Online materials and methods}\label{app:online-methods}

\subsection{The SourceData-NLP dataset}\label{sec:soda_dataset}

The curation workflow takes place on the SourceData platform~\citep{sourcedata}. Curation is carried out by professional curators from Molecular Connections\footnote{https://molecularconnections.com/}. When curators are uncertain, or in some systematically predefined cases, additional information or identifier validation is requested from the authors (see policies in Supplementary Material~\ref{app:guidelines}). In the EMBO Press editorial process, authors are requested to provide their source data for figures showing relevant experimental results in their papers. These figures are then sent to the curation team, to be annotated following the guidelines described below and Supplementary Material~\ref{app:guidelines}\footnote{The guidelines are also maintained at https://sourcedata.embo.org/documentation/}. 


\subsubsection{Annotation process}\label{sec:scope}

 We provide here an overiew of the four major steps of the curation process: splitting figures into panels, tagging of entitites, linking entities to identifiers and, categorization of the role of entities in the experimental design. 

\paragraph{Panel segmentation:} 
In the life sciences, figures commonly include multiple panels that present results obtained through various approaches. For efficient curation and representation of the experimental design, annotation is performed at the level of individual panels. The figure panels serve as a coherent unit of research, often describing results from a single experimental assay and a specific experimental system. The tagging procedures detailed below are applied to all panels, except those that display schematics, computational simulation results, overviews, or workflows.

\paragraph{Entity tagging:}
Specific biochemical terms within a panel legend are tagged and classified into eight mutually exclusive entity classes: small molecules, gene products (genes and proteins), subcellular components, cell types, cell lines, tissues, organisms, and diseases. These classes span various organizational scales in biological organisms, from small molecules to species. Generic references to these classes, such as "proteins," "genes," and "cells," are not included in the annotation process. Post-translational modifications, mutations, and other attributes of an entity are not tagged, focusing solely on the base terms. To enable the description of the empirical design, we also tag the experimental assays as an additional class.

\paragraph{Entity linking:} 
Biological entities and experimental assays are normalized and linked to their corresponding identifiers in ontologies. Normalization ensures the unambiguous identification of entities. In cases where a single identifier cannot be assigned definitively, data curators are instructed to assign multiple possible identifiers to the entity. The ontologies used in SourceData-NLP, listed in Supplementary Table~\ref{tab:ontologies}, are carefully selected to facilitate the annotation process, with preference given to curated entries. General definitions are given below and detailed examples are shown in Supplementary Material~\ref{app:guidelines}.

\paragraph{Entity roles:} 
The entities from the previous step are further categorized based on the experimental design. SourceData-NLP utilizes ``role'' to represent causal hypotheses that are experimentally tested. There are six defined experimental roles, as described in~\citet{sourcedata}: 

\begin{itemize}
    
    \item \textit{Measured variable} --- A measured variable is the component that is measured or observed.

    \item \textit{Controlled variable} --- A controlled variable (also called perturbation, intervention, manipulation, alteration, or independent variable) is a component that is experimentally altered. A controlled variable must be targeted and must be controlled. This implies that the experiment must involve the same experimental system across experimental groups and must involve a comparison between several experimental groups to test whether the controlled variable causes an effect on the measured variable. 
    
  \item \textit{Experimental variable} --- When a component is used to compare multiple experimental groups but it is not possible to infer a cause-and-effect relationship between this component and the measured variables of the experiment, the component is said to be an experimental variable.

     \item \textit{Biological component} --- A biological component is a generic category for any experimentally relevant component that does not fit any of the other defined roles within SourceData. Often it will contain the organism, the cell, or a generic treatment that is present across all conditions.
    \item \textit{Reporter component} --- A reporter component is used as a proxy to measure or observe indirectly a measured variable of interest to which it is linked as part of a synthetic or engineered construct.
    
    \item \textit{Normalizing component} --- A normalizing component is a component that is assayed to provide baseline measurements from each experimental group so that the data can be normalized across groups.
    
  As per definition each experiment must have at a minimum a measured variable. In cases where important biological entities (e.g. ``Measured Variable'' or ``Controlled Variable'') are not explicitly mentioned in the figure legends text, the entities are added as 'floating tags'. Floating tags are, like figure-legend-based entities, linked to the respective identifiers. 
    
\end{itemize}

\paragraph{Non-entity tagging for experimental assays}

In addition to entities, SourceData also tags non-entity terms. In particular, the experimental assay used to observe or measure the measured variables of an experiment is tagged. The experimental assays are normalized to identifiers either from the BioAssay Ontology (BAO) or from the Ontology for Biomedical Investigations (OBI). Other non-entity tags are time-related variables and physical variables. A time-dependent variable such as ``time course'' or ``age'' is added when the experimental design includes a time-dependency. Physical variables refers to particular physical experimental conditions, e.g. \texttt{cold exposure}, \texttt{footshock}, etc.

\begin{table}[!t]
    
    \caption{Ontologies to which the SourceData-NLP tagged entities are normalized.}
    \label{tab:ontologies}
    \centering
    \begin{tabularx}{0.7\textwidth}{l|ll}
        \toprule 
        \textbf{Entity Type}               & \textbf{Primary Resource}           & \textbf{Secondary Resource}      \\
        \midrule
        \textbf{Small molecules}           & ChEBI$^1$                           & PubChem$^{12}$                   \\
        \textbf{Genes}                     & NCBI Gene$^2$                       & Rfam$^{13}$                      \\
        \textbf{Proteins}                  & UniprotKB/Swiss-Prot$^3$            &                                  \\
        \textbf{Subcellular components}    & Gene Ontology$^{4,5}$               &                                  \\
        \textbf{Cell types and cell lines} & Cellosaurus$^6$                     & Cell Ontology$^{14}$             \\
        \textbf{Tissues \& organs}         & Uberon$^7$                          &                                  \\
        \textbf{Organisms \& species}      & NCBI Taxonomy$^{8,9}$               &                                  \\
        \textbf{Diseases}                  & Disease Ontology$^{15}$              & MeSH$^{16}$                      \\
        \textbf{Experimental assays}       & BAO$^{10}$, OBI$^{11}$              &                                  \\
        \bottomrule
    \end{tabularx}
    \par
    \parbox[]{0.7\textwidth}{
        \footnotesize{
            $^1$~\citealp{chebi}, $^2$~\citealp{ncbigene}, $^3$~\citealp{uniprot}, 
            $^4$~\citealp{geneont1}, $^5$~\citealp{genont2}, $^6$~\citealp{cellosaurus}, 
            $^7$~\citealp{uberon}, $^8$~\citealp{ncbitaxonomy1}, $^9$~\citealp{ncbitaxonomy2}, 
            $^{10}$~\citealp{bao}, $^{11}$~\citealp{obi}, $^{12}$~\citealp{pubchem}, $^{13}$~\citealp{rfam}, $^{14}$~\citealp{cellont}, $^{15}$~\citealp{do}, $^{16}$~\citealp{mesh}
        }
    }
\end{table}

\subsubsection{Validation and quality control}

The SourceData annotation workflow includes a quality control step in which a second annotator performs spot checks on the annotations. This process helps identify and rectify inconsistencies in the labeling of biological entities that are inevitably introduced by human annotators during the workflow. Since the annotation process spanned several years, the guidelines were continuously refined to address systematic and common annotation errors.

To address systematic errors outside the workflow, the following corrections were implemented:
\begin{itemize}
    \item[1.] Exclusion of panels lacking at least one ``measured variable'' (the entity under study).
    \item[2.] Verification of entities with different IDs but identical text to ensure that differences were not due to annotation errors.
    \item[3.] Compilation of a list of overly generic terms (e.g., ``image,'' ``percent,'' or ``cell'' – see the complete list in Appendix~\ref{app:generic-terms}) that annotators commonly used, and removal of annotations associated with those terms.
\end{itemize}

These measures have improved the accuracy of our annotations. However, it is important to emphasize that the effectiveness of these corrections ultimately relies on the expertise and vigilance of our human annotators in identifying patterns and inconsistencies.
\newpage
\section{SourceData guidelines\label{app:guidelines}}
\subsection{Introduction}\label{app:introduction}

Experiments in cell and molecular biology involve the empirical manipulation, observation, and description of biological entities. Biological and chemical entities can be entire organisms, a subset of their constituents, or part of the experimental milieu.

\begin{note}
    In this document, the terms entity and component are used interchangeably.
\end{note}

\begin{example}
    Calcium, oligomycin, p53, mitochondria, liver, mus musculus, synapse, and HeLa cells are entities.
\end{example}

\begin{example}
    The cell cycle, apoptosis, wound healing, or type II diabetes are not entities.
\end{example}

SourceData description of the data presented in scientific figures specifies the entities that are relevant to the scientific meaning of the data. Annotation of attributes of such entities, biological processes, or diseases is not yet part of the SourceData specification described in this document.

In the following sections of this document, we define the key concepts used in the SourceData annotation process, including the partitioning of composite figures into coherent panels, the tagging of entities, their assignments to types and roles, and their normalization using external identifiers.

\subsection{Partitioning figures into panels}\label{app:partitioning-figures-into-panels}

Conventional figures are composed of multiple panels and are associated with a description, the figure legend (or figure caption), that explains the content of the figure. While figures tend to present a heterogeneous mixture of experimental designs and assays, individual panels are much more coherent. SourceData annotation is therefore carried out at the level of individual panels.

A panel should be defined as a subset of a full figure such that all of the data points/measurements/observations included in the panel are comparable to each other in a scientifically meaningful way. It is often possible to define a single common observational assay across all observations/data points presented in the same panel. In the majority of cases, panels correspond to the visual panels spontaneously delimited by authors.

Each panel must be associated with its specific panel legend.

\begin{note}
    It is often possible to generate a panel legend by including the appropriate textual fragments of the full figure legend. In some instances, multiple non-contiguous fragments need to be spliced together.
\end{note}

\subsection{Tagging entities}\label{app:tagging-entities}

The primary source of information for SourceData annotation is the text of the panel legend and the image of the figure. Relevant terms from the legend or from the image are attached to a tag that specifies their type and role and that can be further linked to identifiers from external biological databases.

\subsubsection{Tagging terms in figure legends}\label{app:tagging-terms}
To be tagged, a panel must report experimental data. In the text of a panel legend, terms that correspond to specific biological and chemical entities should all be tagged.

\begin{note}
    Panels that present schematics, computational simulation results, overviews, and workflows are not tagged.
\end{note}

\begin{figure*}[ht]
    \centering
    \includegraphics[width=0.75\linewidth]{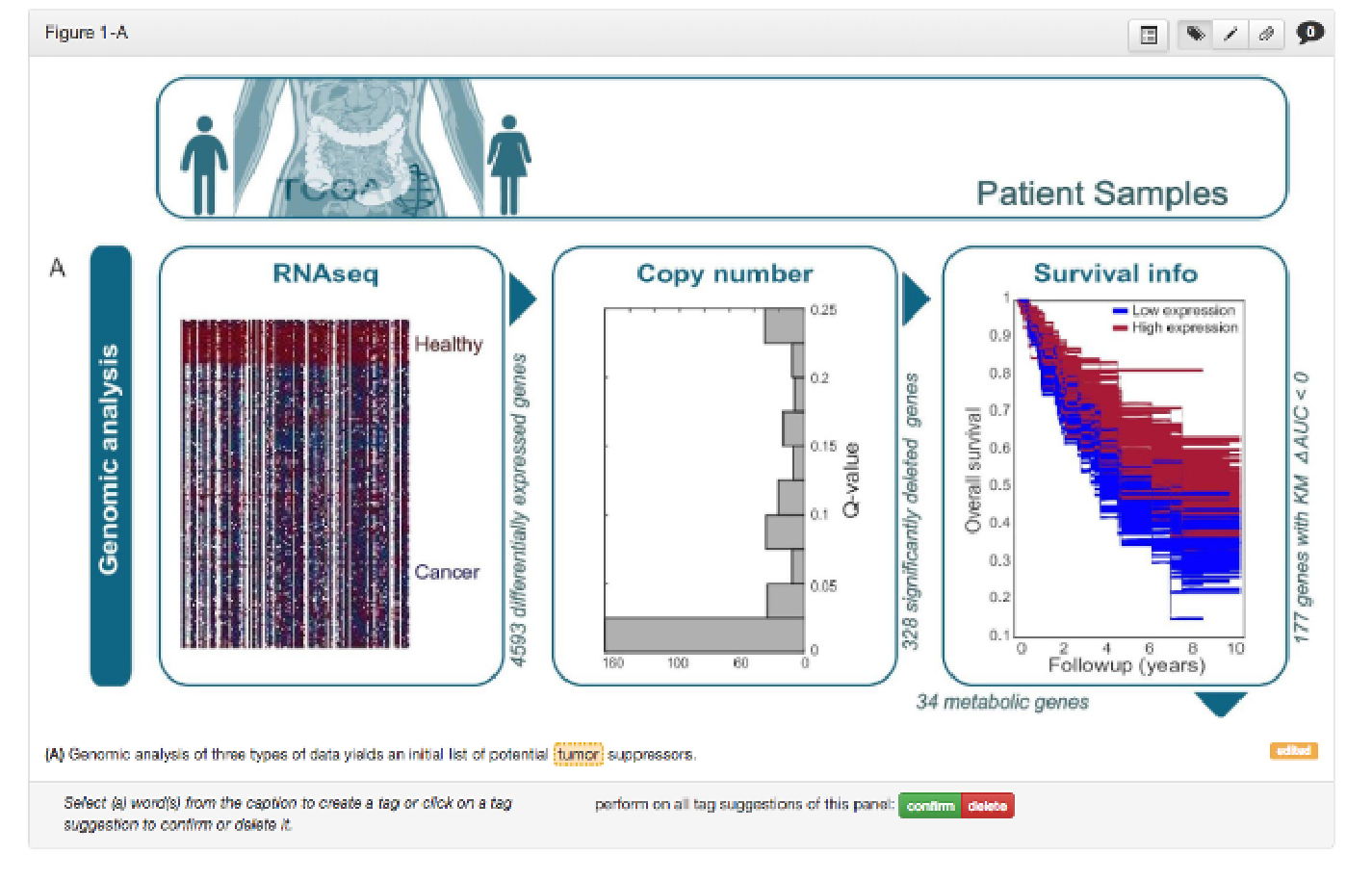}
    \caption{Schemes like the one shown above do not need to be annotated.}
    \label{fig:scheme}
\end{figure*}

In general, generic terms referring to broad classes of biological components (e.g., 'proteins', 'cells', 'animals') should not be tagged unless they refer to the object of an assay.

Some terms such as those referring to proteins or genes can be appended with prefixes or suffixes that indicate a post-translational modification, a mutation or other variations of the actual base term. In such cases, pre- or suffixes should be left out and only the base term should be tagged. In other cases, a prefix is added to an entity to denote a species origin, in which case the prefix should be kept.

\begin{example}
    If the cancer-related mutant form of B-RAF is mentioned as \texttt{B-RAF(V600E)} in the text of the legend, the suffix \texttt{(V600E)} indicating the mutation should be ignored and only \texttt{B-RAF} should be tagged. Similarly, if \texttt{p-Akt1} is designating the phosphorylated form of Akt1, only \texttt{Akt1} should be tagged and the prefix 'p-' should be left out.
\end{example}

\begin{example}
    The protein dMyc refers to the Drosophila Myc protein homolog and should be tagged as \texttt{dMyc}.
\end{example}

\begin{note}
    Some components are engineered by assembling or fusing multiple sub-components, which should be tagged individually. For example, the term \texttt{RAS-GFP} referring to a fusion protein between GFP and RAS should be annotated with two tags: \texttt{RAS} and \texttt{GFP}.
\end{note}

\begin{note}
    In some instances, a fusion construct of multiple entities can be created and referred to within the text via a shorthand symbol generated by authors. In this case, floating tags should be created to refer to the individual components of the fusion construct rather than tagging the shorthand symbol as an entity.
\end{note}

\subsubsection{Adding terms missing from the figure legend}\label{app:adding-terms}

Terms can be added as floating tags to complement the description of an experiment with entities that are missing from the text of the legend. These entities typically appear in the image of the figure but not in the legend.

The use of floating tags should be restricted only to entities with role \texttt{controlled variable}, \texttt{measured variable}, or \texttt{experimental variable}. If one or several of the 3 elements are missing, they should be annotated as floating tags.

\begin{figure*}[ht]
    \centering
    \includegraphics[width=0.75\linewidth]{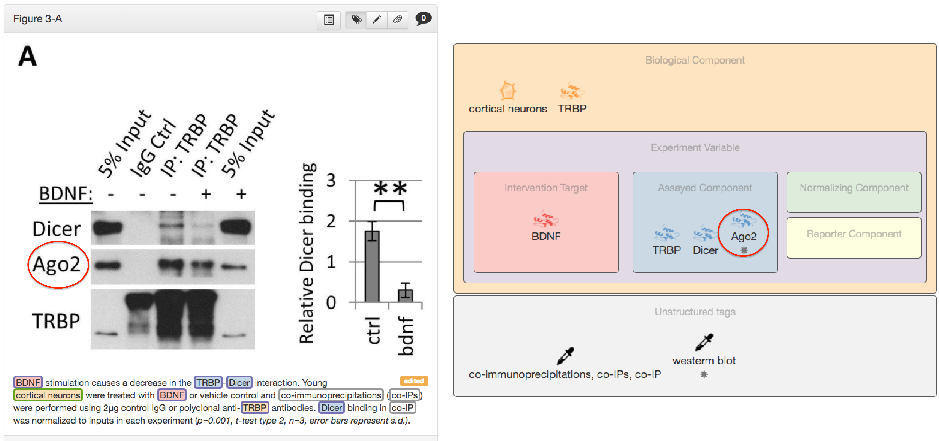}
    \caption{Key elements missing in the figure caption should be added as a floating tag: in this example, the measured variable component AGO2 is missing and was added as a floating tag.}
    \label{fig:floating_tag}
\end{figure*}

When a generic term is used, such as \texttt{cell} or \texttt{transcripts}, to refer to a specific entity, i.e., a specific cell line or a specific mRNA, a floating tag should be used to refer explicitly to the specific entity.

\subsection{Entity types \& normalization}\label{app:entity-types-normalization}

\subsubsection{Types}\label{app:types}

Entities are assigned to one of seven types spanning successive levels of biological organizations (~\ref{tab:kb}). Each type is mutually exclusive. If an entity is linked to an identifier from an external resource, it should use the resource associated with its type according to ~\ref{tab:kb}.

\begin{example}
    \texttt{ATP} is a small molecule, \texttt{creb1} is a gene, \texttt{CREB1} is a protein, \texttt{the Golgi apparatus} is a cellular component, \texttt{HEK293} is a cell line, the \texttt{retina} is a tissue, and \texttt{Saccharomyces cerevisiae} and \texttt{PhiX174} are organisms.
\end{example}

If an entity does not fit any of the predefined types, the undefined type is assigned. By definition, undefined entities cannot be linked to any external resources.

\begin{note}
    Undefined entities are tagged as such to enable a retrospective analysis of whether additional resources and types should be added in a subsequent version of the SourceData model.
\end{note}

In omics experiments, the number of entities measured is too large to be all listed explicitly. It is then possible to represent the experiments by adding as a floating tag the reserved words \texttt{multiple components} and assigning the appropriate type.

\subsubsection{Linking to standard identifiers (normalization)}\label{app:linking}

In the normalization process, entities should be linked to one or several identifiers of the external resources corresponding to the entity's type (see~\ref{tab:kb}). If an entity is linked to multiple identifiers, it must mean that there is uncertainty about the exact identity of the entity.

\begin{example}
    If the term \texttt{Akt} is used to refer to the mouse protein Akt, it is unclear whether it refers to the Akt1, Akt2, or Akt3 isoform. As such, the term will be normalized to the external identifiers \texttt{Uniprot:P31750; Uniprot:Q60823; Uniprot:Q9WUA6} corresponding to Akt1, Akt2, and Akt3, respectively.
\end{example}

\begin{note}
    In the case of entities that are normalized to identifiers from ontologies and taxonomies (subcellular components, cell types, tissues, and organisms), uncertainty about the identity of the entity should be expressed by normalizing it to a sufficiently generic concept in the ontology/taxonomy. For example, the strain \texttt{HSV-1 (F)} does not have a specific entry in the NCBI Taxonomy database but can be normalized to the more generic taxon \texttt{HSV-1[NCBI Taxonomy:10304]}.
\end{note}

Linking reporter components or normalizing components to an external identifier is optional.

Identifiers pointing to curated records of external databases should be preferred over identifiers referring to non-curated records. If relevant records exist both in the primary and secondary resources listed in ~\ref{tab:kb}, identifiers from the primary resource should be used.

\subsection{Entity roles}\label{app:entity-roles}

Biological components listed in the caption of a figure each play a different role in the experimental design: some components are altered in a controlled manner, others remain untouched by the experimenter, and some are directly or indirectly measured to perform measurements or observations. Accordingly, the following roles are defined:
\begin{itemize}
    \item Biological component -A biological component is a generic category for any experimentally relevant component that does not fit any of the other defined roles within SourceData. Often it will contain the organism, the cell, or a generic treatment that is present across all conditions.

    \item Measured variable - 
    \item Controlled variable
    \item Reporter component
    \item Normalizing component
    \item Experimental variable
\end{itemize}

\begin{note}
    For all types of entities, if there are multiple instances of the same entity in the figure legend, all instances of the tag should be captured.
\end{note}

\subsubsection{Biological components}\label{app:biological-components}
A biological component is a generic category for any experimentally relevant component that does not fit any of the other defined roles within SourceData. Often it will contain the organism, the cell, or a generic treatment that is present across all conditions.

\subsubsection{Measured variables}\label{app:assayed-components}
A measured variable is the component that is measured or observed.

\begin{example}
    The proteins detected on a Western blot are the measured variables except the loading control, if any, which is considered as a normalizing component (see below).
\end{example}

\begin{figure*}
    \centering
    \includegraphics[width=0.75\linewidth]{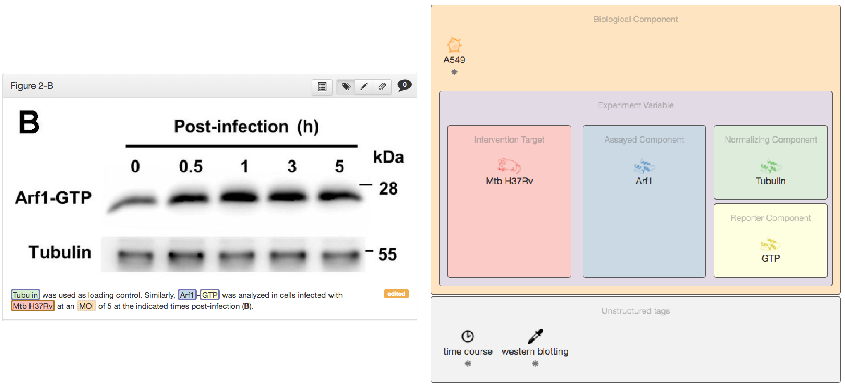}
    \caption{In the example above, tubulin is the normalizing component.}
    \label{fig:normalizing_western}
\end{figure*}

\begin{note}
    If a molecular marker, for example, the protein \texttt{EEA1}, is measured to visualize a higher order structure, for example, endosomal vesicles, the marker (EEA1 in this example) is tagged as a measured variable. The higher-order structure (endosomal vesicles in this example) is tagged as a measured variable only if it is explicitly highlighted on the image or a property of the entity (such as number/localization) is mentioned in the text of the legend.
\end{note}

\begin{figure*}
    \centering
    \includegraphics[width=0.75\linewidth]{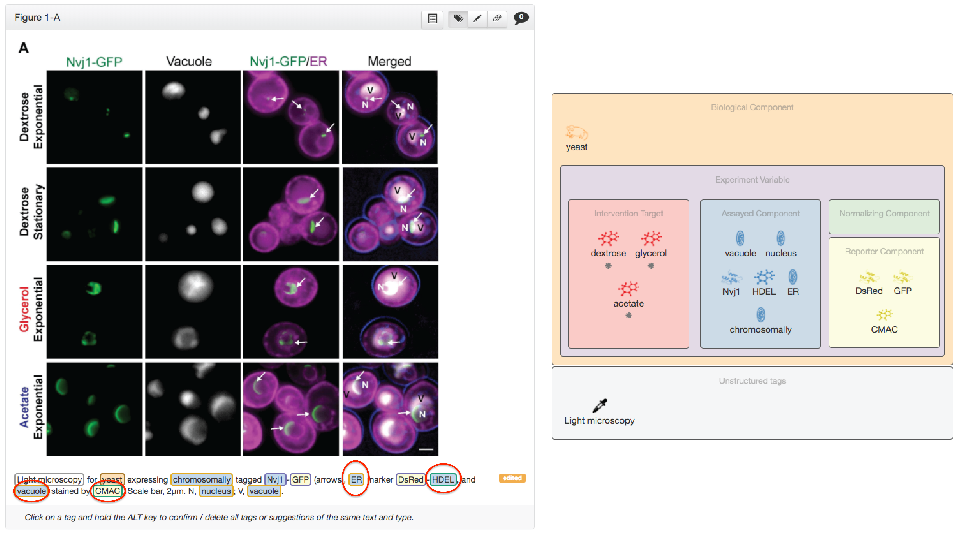}
    \caption{In this case, \texttt{HDEL} is an ER marker, so it is captured as a measured variable. In addition, because there is a specific reference to the ER in the legend, ER is also captured as a measured variable. Note that in this example there is a second marker, \texttt{CMAC}, which is, however, a reporter as is therefore captured as a reporter for vacuole, which is labeled as a measured variable.}
    \label{fig:markers}
\end{figure*}

\subsubsection{Controlled variables}\label{app:controlled-variables}
A controlled variable (also called perturbation, intervention, manipulation, alteration, or independent variable) is a component that is experimentally altered. A controlled variable must be targeted and must be controlled. This implies that the experiment must involve the same experimental system across experimental groups and must involve a comparison between several experimental groups to test whether the controlled variable causes an effect on the measured variable.

\begin{example}
    The function of the gene creb1 can be investigated by comparing creb1 wt (control group) to creb1-/- knockout (test group) mice; in this experiment, creb1 is the controlled variable. If, and only if, it is appropriately controlled, the purpose of such an experiment is to infer a cause-and-effect relationship, whether direct or indirect, between the controlled variable and the measured variable.
\end{example}

\begin{warning}
    If a drug (cycloheximide, for example) is applied across all experimental groups, it is not considered a controlled variable, since there is no control group to compare the effect of the drug across conditions. A controlled variable must be controlled. Accordingly, in such a context, the drug should be tagged as a biological component. Similarly, if a cell strain harboring the same genetic mutation is used across all experimental groups, the mutated gene is not a controlled variable but a generic biological component.
\end{warning}

\begin{figure*}
    \centering
    \includegraphics[width=0.75\linewidth]{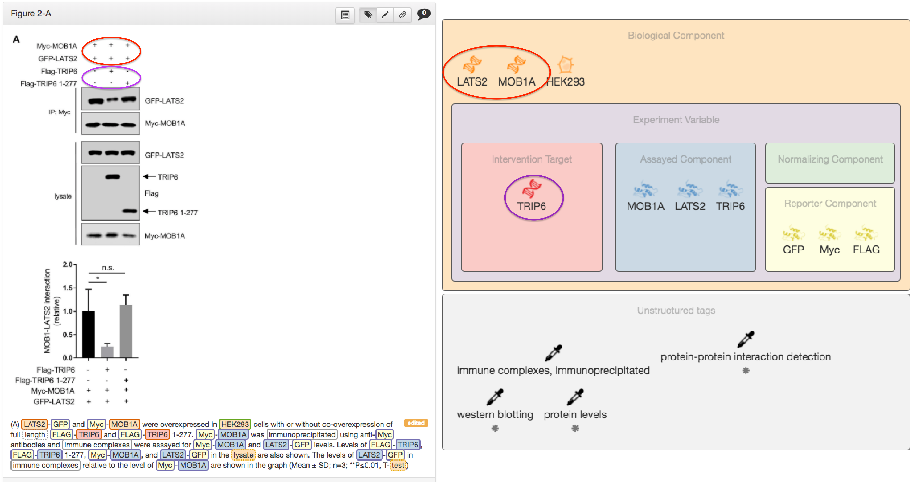}
    \caption{\texttt{MOB1A} and \texttt{LATS2} are both overexpressed across all conditions in this experiment, i.e., they are not controlled for. The only entity that is differentially manipulated in this experiment is \texttt{TRIP6}. Thus, both \texttt{MOB1A} and \texttt{LATS2} are considered biological components and \texttt{TRIP6} is considered the controlled variable.}
    \label{fig:biological_component}
\end{figure*}

\begin{note}
    The target of an experimental manipulation is usually tagged as a controlled variable. Small molecules such as drugs, inhibitors, agonists, and other pharmacological compounds are usually considered as the controlled variable when their effects are compared across experimental groups. An exception is when a small molecule (for example, doxycycline, IPTG, arabinose) is used to manipulate the activity of an engineered circuit controlling the actual entity of interest (for example, a gene whose expression needs to be varied), in which case the entity of interest is considered as the controlled variable and the triggering compound (doxycycline, IPTG, arabinose) is captured as a biological component.
\end{note}

\begin{example}
    If cells are treated with different doses of the PKA inhibitor \texttt{H89}, \texttt{H89} is tagged as the controlled variable. 
\end{example}

\begin{note}
    In experiments that test the action of an entity over time, the entity is tagged as a controlled variable only if a control group is tested or if the \texttt{time = 0} is also shown as a point of comparison.
\end{note}

\begin{example}
    In a siRNA-mediated knockdown experiment, the gene targeted by the siRNA is tagged as a controlled variable.
\end{example}

\begin{note}
    A controlled variable \textit{must} involve controlled experimental conditions. It is therefore common that control experimental groups are treated with a neutral compound, for example, the solvent used to dissolve the administered drug. By convention, such components MUST be assigned the generic role of biological components.
    
    In transfection experiments for overexpression, the main entity of the construct used for transfection should be labeled as a controlled variable of type gene and if detected, the protein should be tagged as a measured variable of type protein.
\end{note}

\subsubsection{Reporter components}\label{app:reporter-components}
A reporter component is used as a proxy to measure or observe indirectly a measured variable of interest to which it is linked as part of a synthetic or engineered construct.

\begin{example}
    A RAS-GFP fusion protein includes the RAS protein as a measured variable and GFP as a reporter component.
\end{example}

\begin{example}
    The luciferase gene can be used as a reporter gene to monitor the transcriptional activity of a given gene promoter, which is the actual measured variable of interest.
\end{example}

\begin{figure*}
    \centering
    \includegraphics[width=0.75\linewidth]{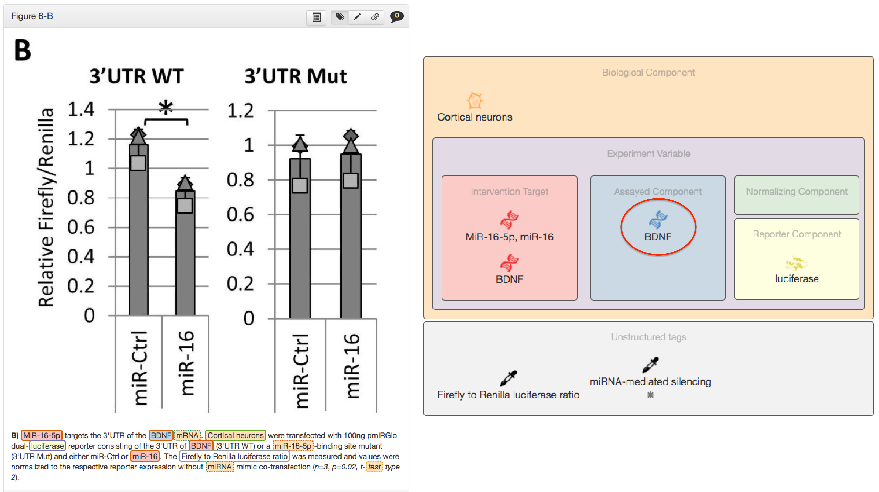}
    \caption{In this luciferase experiment, what is being measured is the effect of miR-16-5p on the 3'UTR of Bdnf, so \texttt{Bdnf} is labeled as the measured variable.}
    \label{fig:luciferase}
\end{figure*}

Linking reporter components or normalizing components to an external identifier is optional.

\subsubsection{Normalizing components}\label{app:normalizing-components}
A normalizing component is a component that is assayed to provide baseline measurements from each experimental group so that the data can be normalized across groups.

\begin{example}
    The proteins beta-actin or GAPDH are often assayed to serve as loading control in Western blots and are then tagged as normalizing components.
\end{example}

Linking normalizing components to an external identifier is optional.

\subsubsection{Experimental variables}\label{app:experimental-variables}
When a component is used to compare multiple experimental groups but it is not possible to infer a cause-and-effect relationship between this component and the measured variables of the experiment, the component is said to be an experimental variable.

\begin{example}
    If the expression of a given gene is measured across tissues and cell lines, including liver, muscle, brain, HEK293, and HeLa cells, the tissues or cell types are tagged as experimental variables.
\end{example}

\begin{figure*}
    \centering
    \includegraphics[width=0.75\linewidth]{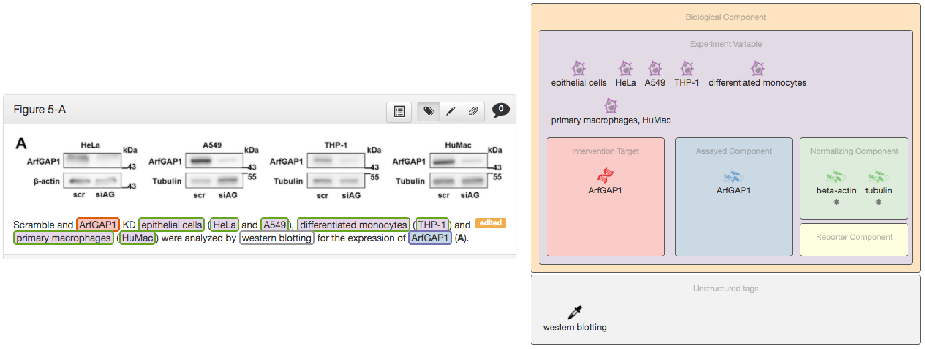}
    \caption{In this example, the expression of \texttt{ArfGAP1} is assayed in several cell types in parallel. This is the prototypical definition of an experimental variable.}
    \label{fig:experimental_variable}
\end{figure*}

\subsection{Tagging experimental assays}\label{app:tagging-experimental-assays}
In addition to entities, SourceData is also experimenting with the tagging of non-entity terms. In particular, the experimental assay used to observe or measure the measured variables of an experiment is tagged and normalized to identifiers either from the BioAssay Ontology (BAO) or from the Ontology for Biomedical Investigations (OBI).

\begin{note}
    At present, SourceData only captures experimental assays used to collect data, i.e., associated with the measured variable. The experimental assay used to induce the controlled variable should not be tagged. In addition, SourceData strives to capture the experimental assay itself and not necessarily the METHOD. If both the assay and the method are explicitly mentioned in the figure legend, SourceData captures both. If not, only the experimental assay is captured, either as a tag in the figure legend or as a floating tag if missing.
\end{note}

\begin{example}
    If a figure represents images from an immunostaining, it suffices to annotate \texttt{immunostaining} as the experimental assay and there is no need to add a floating tag for \texttt{microscopy} if this is not present in the figure legend.
\end{example}

\begin{figure*}
    \centering
    \includegraphics[width=0.75\linewidth]{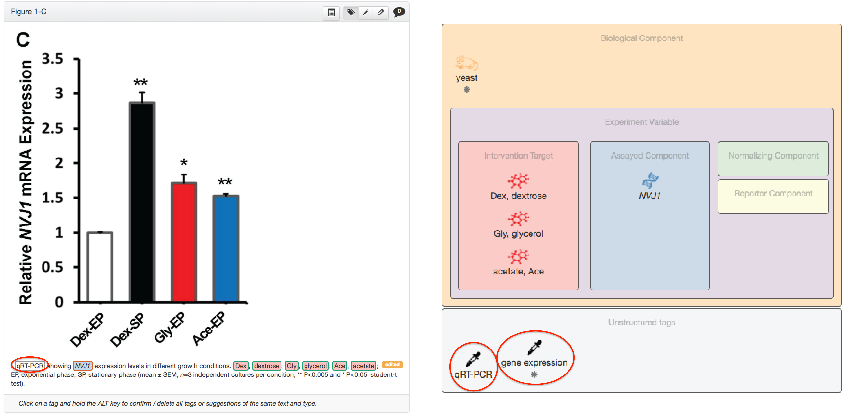}
    \caption{In this example, the experimental assay is \texttt{gene expression measurement} and the method is qRT-PCR. Because qRT-PCR is mentioned in the figure legend, it is captured. However, \texttt{gene expression} is also added as a floating tag to capture the experimental assay for the experiment.}
    \label{fig:assay_method}
\end{figure*}

\subsection{Time-related variables}\label{app:time-related-variables}
Add a floating tag \texttt{time course} or \texttt{age} to indicate a comparison of a controlled variable or a measured variable over time within an experiment.

\subsection{Physical variables}\label{app:physical-variables}
A Physical variable refers to particular experimental conditions, e.g. \texttt{cold exposure}, \texttt{footshock}, etc. Add them only when explicitly mentioned in the figure legend.

\subsection{Special cases: Protein complexes, FACS, cell cycle phases, and DNA staining}\label{sec:special-cases}
Some experimental designs are unique. For the following cases, these guidelines should be observed:

\begin{itemize}
    \item Protein complexes: although the intuitive normalization for protein complexes would be of type protein, it is more adequate to assign them the type \texttt{subcellular component} because the Gene Ontology (GO) database contains normalized references for protein complexes. An example of this would be RNA Polymerase II, which is made up of a number of individual subunits.
    
    \begin{figure*}
    \centering
    \includegraphics[width=0.75\linewidth]{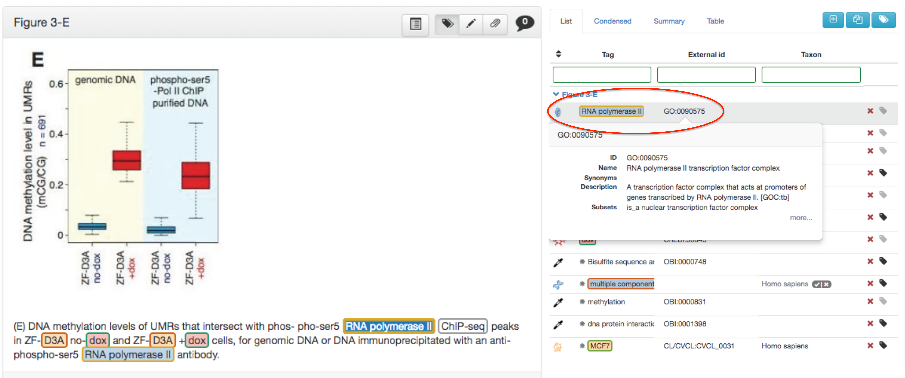}
        \caption{Notice how in this example RNA-Pol-II is assigned the type \texttt{subcellular component}, which can be normalized to the GO database.}
        \label{fig:protein_complexes}
    \end{figure*}

    \item FACS experiments: for FACS experiments, SourceData captures both the cells and the DNA (or whichever element is stained and sorted) as measured variables.
    
    \begin{figure*}
    \centering
    \includegraphics[width=0.75\linewidth]{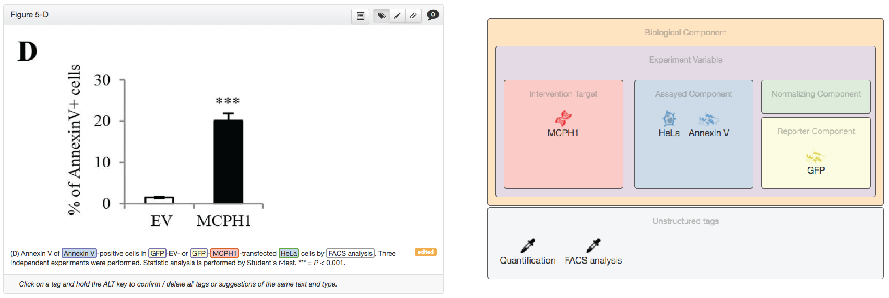}
        \caption{In this example, HeLa cells are stained with AnnexinV to measure cell death. Both HeLa cells and AnnexinV are captured as measured variables.}
        \label{fig:FACS}
    \end{figure*}

    \item Cell cycle phases: at present, SourceData does not include cell cycle phases as time elements. Cell cycle phases may be added when appropriate to experimental variables or biological components as the class \texttt{unknown}.
    \item DNA staining: for DNA stains like BrdU, EdU, etc., SourceData captures DNA as the measured variable and \texttt{BrdU staining} as the experimental assay.
\end{itemize}

\subsection{Representing 'omics' experiments}\label{app:representing-omics-experiments}
For experiments performing a large number of measurements (>15-20), for example in metabolomics, genomics, transcriptomics, and proteomics, the measured variables cannot be listed individually. The following tentative guidelines are then followed:

\begin{itemize}
    \item The reserved expression \texttt{'multiple components'} should be included as a floating tag, with the relevant entity type, and with the role measured variable.
    
    \begin{figure*}
        \centering
        \includegraphics[width=0.75\linewidth]{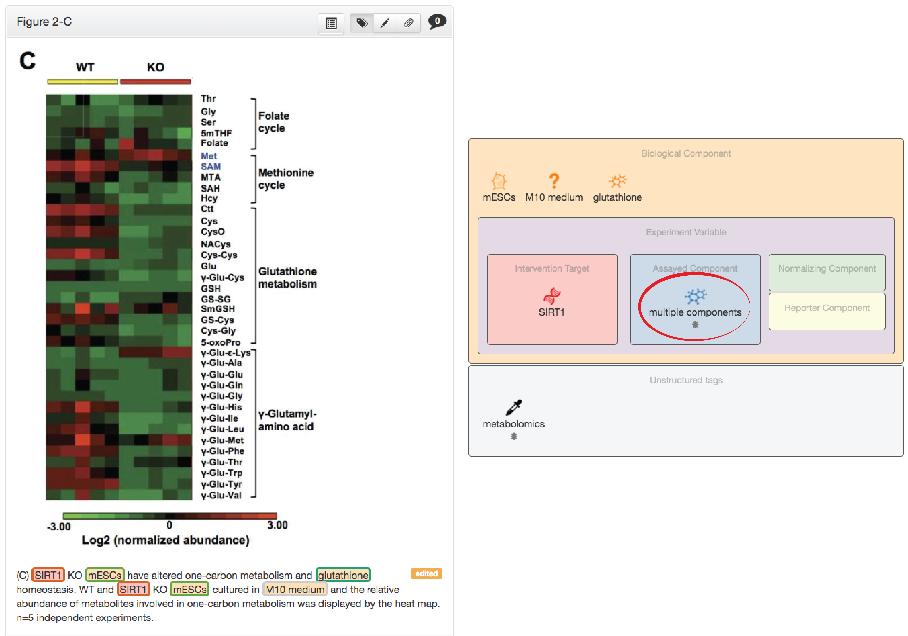}
        \caption{Note how individual metabolites are captured under the expression \texttt{multiple components} in this example.}
        \label{fig:multiple_components}
    \end{figure*}

    \item Both the measurement type (protein expression, protein-DNA interactions, protein-protein interactions, etc...) and the assay technology (experimental platform such as the sequencing platform, mass spec platform, etc...) as mentioned in Materials \& Methods should be tagged as experimental assay, if necessary as floating tag.
    \item The experimental system that is profiled should be added as a biological component, even if it requires adding a floating tag.
\end{itemize}

\begin{example}
    In proteomics, if mass spectrometry is used to measure protein-protein interactions in HeLa cell extracts, for example, the terms \texttt{mass spectrometry} (BAO:0000055) and \texttt{protein-protein interaction assay} (BAO:0002990) will both be added as experimental assay floating tags, as well as \texttt{HeLa} (CL/CVCL:CVCL\_0030) as a general biological component that specifies the experimental system.
\end{example}

\subsection{Guidelines for author queries}\label{app:guidelines-for-author-queries}
When in doubt about the normalization of an entity, authors can be queried via the validation interface. The following should be systematically queried to ensure accurate annotation: disease, cell lines, strains, cell types, and molecules if these are normalized to PubChem.

\newpage
\section{Benchmarking of NLP tasks using SourceData-NLP}\label{sec:nlp-benchmark}

To highlight the usefulness of SourceData-NLP, we conducted a set of experiments to train and benchmark the best existing language models across the different tasks supported by our dataset. Previous works have consistently shown that biomedical language models such as PubMedBERT~\citep{pubmedbert}, and BioLinkBERT ~\citep{biolinkbert} that use a biomedical vocabulary (i.e., where the tokenization model was trained only on biomedical texts) outperform both biomedical language models that use a general vocabulary (BioBERT~\citealp{biobert}; BioMegatron~\citealp{biomegatron}; BioMedRoBERTa~\citealp{biomedroberta}) and general domain language models (BERT~\citealp{bert}; RoBERTa~\citealp{roberta}). Accordingly, the experiments of this work were carried out with the PubMedBERT and BioLinkBERT models using both their large and base versions.

We perform the benchmarking on four tasks: splitting the caption into panels ("panel segmentation"), NER, and the novel semantic interpretation of empirical roles. The best-performing model for each task is available on the EMBO repository hosted on the HuggingFace transformers Hub\footnote{https://huggingface.co/EMBO}. We next describe the fine-tuning procedures in detail.

\subsection{Panel segmentation, NER, and semantic interpretation of empirical roles}\label{methods:ner-finetuning}
All the tasks are formulated as token classification problems. In this configuration, the language models are fine-tuned to assign labels to individual tokens. Labels are assigned based on the IOB schema~\citep{biotagging}, which denotes the (B)eginning tokens of entities, tokens (I)nside entities, and tokens (O)ut of entities.

For the panel segmentation task, the entire figure caption is used as input for the model. Tokens are labeled as {\verb'O'} or {\verb'B-PANEL_START'} to indicate the starting position of each panel. The results of the panelization benchmark are shown in the Supplementary material~\ref{sec:results-panelization}.

The tasks of NER and semantic interpretation of empirical roles utilize the text of distinct panels as input. Subsequently, tokens are categorized into one of eight distinct and non-overlapping classes. These include \verb'O', signifying no biological entity of interest; \verb'B-GENEPROD' and \verb'I-GENEPROD', representing gene products; and additional classifications for other biological entities like subcellular components, cell types, cell lines, tissues, organisms, diseases, and experimental assays, as annotated in SourceData-NLP.

The task of assigning empirical roles to the entities is specifically applied to gene products as they are by far the most prevalent entities with these roles. We approach this task in three different ways, as explained in the main text. First, we make the semantic classification task purely dependent on context, masking the gene product entities with the special {\verb'[MASK]'} token and assigning the labels {\verb'CONTROLLED_VAR'}, {\verb'MEASURED_VAR'}, or {\verb'O'} to them (``context-only approach''). Secondly, we mark the position of entities with a special token without masking them (``marked-entity approach''). In this case, the models can learn both from context and from the identity of the entities. Finally, we train the models to recognize the roles of gene products without any extra information, forcing the models to perform both, NER and role interpretation on a single step (``single-step approach'').

To ensure reproducibility, we maintained the same fine-tuning hyperparameters for all tasks as summarized in the Supp. Table~\ref{tab:hyperparam}.  Model parameters were optimized with AdamW with the cross-entropy loss. To avoid memory issues the large version of the models were trained using a batch size of 8 compared to 32 for the base model. To achieve comparable results between both model sizes, we used the same learning rate and 4 gradient accumulation steps for training large models. The reported results are the average F1 scores obtained from 5 consecutive runs of the experiments using different random seeds.

Training is carried out on an NVIDIA DGX Station with 4 Tesla V100 GPUs for 2 epochs. Each fine-tuning takes about 20 minutes for a base model and about an hour for a large model, which has approximately 3 times more parameters. 

\subsection{Multimodal segmentation of compound figures}\label{app:segmentation}

The training of the object detection algorithm in YOLOv10 was done using default parameters. The model was fine-tuned for 30 epochs. The process took 1.5 hours in our NVIDIA DGX Station with 4 Tesla V100 GPUs. The training batch size was automatically adjusted by the algorithm. 

The model evaluation was run on the test set. For the evaluation we set several parameters to maximize the model performance. We set the parameter image size (\verb|imgsz|) to 420, the confidence parameter (\verb|conf|) to 0.3, and the intersection over union parameter (\verb|iou|) to 0.9. With this configuration, the YOLOv10-based object detection algorithm achieves accuracies of $mAP_{50}=98.2\%$ and a $mAP_{50-95} = 87.0\%$.

We note that the panel segmentation of this task was done different from that described in appendix~\ref{methods:ner-finetuning}. In this case, the panelization of the figure caption was performed using API calls to OpenAI GPT multimodal model, GPT-4o. The API calls contained the text of a full figure caption and the image of one of the panels of the correspondent figure extracted with our compound figure segmentation model described above. The return of the model is a panel description that can be understood on its own, without the need of relying on the context of the entire figure caption. The system prompt used for the AI-agent is shown below:

\begin{systemprompt}

    \begin{lstlisting}
        """You will receive a text with the caption of a scientific figure. This figure will be generally composed of several panels. Extract the relevant part of the figure caption so that it matches the panel given as an image file. If a generic description of several panels is in place, return the generic and the specific descriptions for a given panel.  Make sure that the information in the panel caption you return is enough to interpret the panel. For simplicity in post-processing begin the caption always with 'Panel X:' where X is the label of the panel in the figure.
        
        Output format:
        ```
        {
            "panel_label": "X",
            "panel_caption": "Description of the panel."
        }
        ```
        """
        \end{lstlisting}
\end{systemprompt}

We make the source code used for this task openly available for researchers in order to help accelerating research and to reproducing our results. It can be found in \verb|https://github.com/source-data/soda_image_segmentation|.

Combining the two steps we obtain a figure segmentation and panel matching accuracy of 97.7\%, that translates into a single panel miss every six papers, assuming an average of 20 panels per paper, as is extracted from our SourceData-NLP dataset.




\begin{table}
    \centering
    \caption{Hyper-parameters used for fine-tuning.}
    \label{tab:hyperparam}
    \begin{tabularx}{0.5\textwidth}{ll}
        \toprule
        Parameter search space           & Fixed fine-tune param.\\
        \midrule
        train batch size                 & 32 (8)               \\
        eval batch size                  & 64                   \\
        epochs                           & 2                    \\
        gradient accumulation steps      & 1 (4)                \\
        learning rate                    & 1e-4                 \\
        lr scheduler                     & cosine               \\
        adam $\beta_1$                   & 0.9                  \\
        adam $\beta_2$                   & 0.999                \\
        adam $\epsilon$                  & 1e-10                \\
        weight\_decay                    & 0.0                  \\
        adafactor                        & True                 \\
        \bottomrule
    \end{tabularx}
    \par
    \parbox[]{0.5\textwidth}{
        \footnotesize{Note: The numbers in parenthesis show different values for large (>300M parameters) models.}
    }
\end{table}

\newpage

\section{Removal of generic terms from the annotations}\label{app:generic-terms}

During our examination of the assembled dataset, we noticed that numerous general terms —such as age, DNA, cells, and animals— had been tagged. We chose to omit a subset of these non-specific terms from the dataset. This decision was motivated because these terms are often too vague or encompassing. As such, they are not ideal for generating precise descriptions of experimental setups. These are explicitly marked to guide future annotation efforts by the dataset curators. Additionally, we have created a correction patch that eliminates all tags corresponding to these identified terms, thereby ensuring dataset consistency. A comprehensive, alphabetically ordered list of such terms is shown below. These changes will be available in the version 2.0.2 of SourceData-NLP.

\paragraph{List of omitted non-specific terms:} age, aggregates, all, amino acids, analysis, and, animal, animals, antibiotics, area, assay, bacteria, bacterial, based, body weight, cell death, cell, cells, cellular, chromatin, cleaved, complex, concentration, contralateral, core, count, counts, cytoplasmic, cytosolic, dark, distance, DNA, duct, ductal, ducts, embryo, embryos, female, females, fetal, fetus, fibers, fluorescence intensity, fluorescence, fluorescent, gas, gland, glands, hand, heatmap, IHC, images, images, individual, intensity, ipsilateral, laser, length, level, levels, light, line, littermates, lung tumor, male, males, membrane, micrograph, micrographs, morphology, muscle, muscles, nuclear, number, numbers, parasite, parasites, patient, patients, percent, percentage, percentages phenotype, phenotypes, photograph, photographs, plant,  plants, primary cell,  primary cells, protein, proteins, pulse, puncta, punctae, quantification, ratio, rest, RNA, SEM, sequence, size, stained, staining, stem, strain, tailed, TEM, temperature, test, tests, the, tumor, tumors, virus, viruses, weeks, weight, white.

\newpage
\section{Additional Results}




\subsection{Panel segmentation task}\label{sec:results-panelization}

The panel segmentation task splits figure captions into their constituent panels by assigning a \verb|B-PANEL_START| label to the first token of each panel. We fine-tuned BioLinkBERT and PubMedBERT to evaluate how well these models would perform. Both the models perform similarly as shown in Table~\ref{tab:panelization-results}. We also studied how well they could memorize versus generalize for this task. Panel separators in the biomedical literature are not very varied and tend to be consistent (e.g. A, a),  (A), (a)), so most examples were considered "memorized." "Generalized" examples were figures without explicit panel separators, where any word might indicate a new panel. This typically happens with content from publishers who do not include the panel separators in their XML content but only as style CSS classes on the live website. Some high-order panel separators (e.g. Z, O) also required generalization.

The models' overall F1 scores were similar for memorized and generalized examples, as expected given that memorized examples made up over 95\% of the data. However, for generalized examples, F1 scores were ~4 points higher on average. These encouraging results show that these language models can distinguish panels based on context alone.

\begin{table}
    \caption{Results of the panel segmentation task for overall, memorized tokens and non-memorized (generalization) tokens.}
    \label{tab:panelization-results}
    \centering
    \begin{tabularx}{0.6\textwidth}{llrrr}
        \toprule
        \textbf{Pretrained model} & \textbf{Size} & \textbf{F1 overall} & \textbf{F1 memo.} & \textbf{F1 gen.} \\
        \midrule
        PubMedBERT  & base  & 90.5          & 90.1          & 94.3          \\
        PubMedBERT  & large & 91.5          & 91.2          & 95.2          \\
        BioLinkBERT & base  & 90.8          & 89.3          & 95.7          \\
        BioLinkBERT & large & \textbf{92.4} & \textbf{92.0} & \textbf{96.5} \\
        \bottomrule
    \end{tabularx}
    \par
    \parbox[]{0.6\textwidth}{
        \footnotesize{The best-performing model for each case is shown in bold.}
    }
\end{table}

\end{document}